\documentclass[a4paper]{article}
\usepackage{amssymb,graphicx,url,amsmath,subfigure
}
\usepackage{times}
\usepackage{tikz}
\usepackage{soul}
\usepackage{color}
\usepackage{xcolor}
\usepackage{algorithm}
\usepackage{algorithmic}
\usepackage{enumerate}
\usepackage{a4wide}
\usepackage{comment}

\newcommand{\Exp}{\mathbf{E}}

\newcommand{\R}{{\mathbb R}}
\newcommand{\N}{{\mathbb N}}
\newcommand{\C}{{\mathbb C}}
\newcommand{\Z}{{\mathbb Z}}
\newcommand{\cH}{{\cal H}}

\newcommand{\bX}{{\bf X}}
\newcommand{\bY}{{\bf Y}}

\newcommand{\bE}{{\bf E}}

\newcommand{\bF}{{\bf F}}

\newcommand{\bb}{{\bf b}}
\newcommand{\ba}{{\bf a}}
\newcommand{\bc}{{\bf c}}
\newcommand{\bg}{{\bf g}}

\newcommand{\sx}{\Sigma_{\bX\bX}}
\newcommand{\se}{\Sigma_{\bE\bE}}

\newtheorem{Theorem}{Theorem}
\newtheorem{Lemma}{Lemma}
\newtheorem{Definition}{Definition}

\newtheorem{Postulate}{Postulate}

\newtheorem{Corollary}{Corollary}

\definecolor{MyDarkGreen}{rgb}{0.17,0.46,0.25} 
\definecolor{MyDarkRed}{rgb}{0.88,0.22,0.21} 
\definecolor{MyDarkBlue}{rgb}{0.11,0.11,0.70}

\definecolor{lightgray}{gray}{0.85}
\sethlcolor{Dandelion}

\usetikzlibrary{arrows,shapes,plotmarks,positioning}
\usetikzlibrary{decorations.markings}

\tikzset{>=stealth'} 
\tikzstyle{graphnode} = 
   [circle,draw=black,minimum size=22pt,text centered,text
     width=22pt,inner sep=0pt] 
\tikzstyle{var}   =[graphnode,fill=white]
\tikzstyle{vardashed}   =[graphnode,draw=gray,fill=white]
\tikzstyle{obs}   =[graphnode,fill=black,text=white]
\tikzstyle{obsgrey}   =[graphnode,draw=white,fill=lightgray,text=black]
\tikzstyle{par}    =[graphnode,draw=white,fill=red,text=black] 
 \tikzstyle{crucial} =[graphnode,draw=white,fill=yellow,text=black] 
\tikzstyle{fac}   =[rectangle,draw=black,fill=black!25,minimum size=5pt]
\tikzstyle{facprior} =[rectangle,draw=black,fill=black,text=white,minimum size=5pt]
\tikzstyle{edge}  =[draw=white,double=black,very thick,-]
\tikzstyle{blueedge}  =[draw=white,double=blue,very thick,-]
\tikzstyle{rededge}  =[draw=white,double=red,very thick,-]
\tikzstyle{prior} =[rectangle, draw=black, fill=black, minimum size=
5pt, inner sep=0pt]
\tikzstyle{dirprior} = [circle, draw=black, fill=black, minimum
size=5pt, inner sep=0pt]

\date{05 April 2017}

\title{Detecting confounding in multivariate linear models via spectral analysis}

\author{Dominik Janzing and Bernhard Sch\"olkopf} 

\begin{document}

\maketitle

\begin{abstract}
We study a model where one target variable $Y$ is correlated
with  a vector $\bX:=(X_1,\dots,X_d)$ of predictor variables  being potential causes of $Y$.
We describe  a method that infers to what extent the statistical dependences between $\bX$ and $Y$
are due to the influence of $\bX$ on $Y$ and to what
extent due to a hidden common cause 
 (confounder) of $\bX$ and $Y$. 
The method relies on concentration of measure
results for large dimensions $d$ and
an independence assumption stating
that, in the absence of confounding,
 the vector of regression 
coefficients describing the influence of each $\bX$ on $Y$ typically has `generic orientation'
relative to the eigenspaces  of the covariance matrix of $\bX$. 
For the special case of a scalar confounder
 we show that confounding typically spoils this 
generic orientation in a characteristic way 
that can be used to quantitatively
estimate the amount of confounding.
\end{abstract}

\section{Introduction and general model}

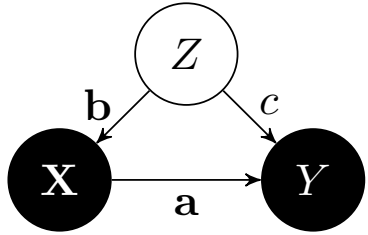
\begin{figure}
\begin{center}
\resizebox{5cm}{!}{
  \begin{tikzpicture}

    \node[obs] at (4,1) (Xc) {$\bX$} ;
    \node[var] at (5,2) (Z) {$Z$} edge[->] (Xc) ;  
    \node[obs] at (6,1) (Yc) {$Y$} edge[<-] (Z) edge[<-] (Xc) ;

    \node[anchor=center] at (5,0.8) {$\ba$ }; 
    \node[anchor=center] at (4.3,1.6) {$\bb$}; 
   \node[anchor=center] at (5.65,1.6) {$c$}; 
  \end{tikzpicture}
}
\end{center}
\caption{\label{fig:generaldag} Most general DAG considered in this paper. All scenarios discussed later refer to this DAG and differ only by the parameters $\ba,\bb,\sigma_F,\se$, $c$.}  
\end{figure}

\begin{figure}
\centering     
\subfigure[]{\label{fig:xydag}}
\resizebox{7cm}{!}{
  \begin{tikzpicture}
    \node[obs] at (4,0) (Xc) {$\bX$} ;
    \node[var] at (5,1) (Z) {$Z$}   ;  
    \node[obs] at (6,0) (Yc) {$Y$} edge[<-] (Z) edge[<-] (Xc) ;
\node[anchor=center] at (7,0) {$\equiv $ };
    \node[obs] at (8,0) (Xa) {$\bX$};
    \node[obs] at (10,0) (Ya) {$Y$}  edge[<-] (Xa);        .
    \node[anchor=center] at (5,-0.2) {$\ba$ }; 
   \node[anchor=center] at (5.65,0.6) {$1$}; 
  \end{tikzpicture}
}
\hspace{1cm}
\subfigure[]{\label{fig:zdag}}
\resizebox{3cm}{!}{
  \begin{tikzpicture}
    \node[obs] at (4,0) (Xc) {$\bX$} ;
    \node[var] at (5,1) (Z) {$Z$} edge[->] (Xc) ;  
    \node[obs] at (6,0) (Yc) {$Y$} edge[<-] (Z) ;
    \node[anchor=center] at (4.3,0.6) {$\bb$}; 
   \node[anchor=center] at (5.65,0.6) {$1$}; 
  \end{tikzpicture}
}
\caption{(a) Purely causal case: setting $\bb=0$
renders $Z$ independent. Thus, we can consider $\tilde{F}:=Z+F$ formally as noise term and then obtain
the simple DAG with nodes $\bX,Y$ only.
 (b) Purely confounded case: by setting $\ba=0$ 
 the correlations between $\bX$ and $Y$ are only due to the confounder $Z$.}
\end{figure}
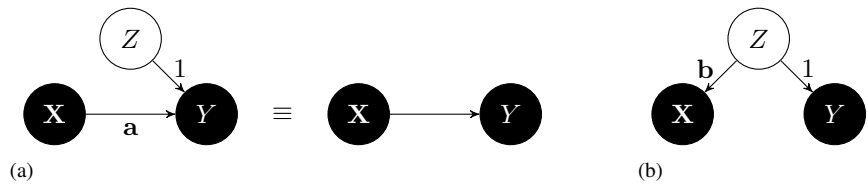

Estimating the causal influence of
some variables $X_1,\dots,X_d$  on a 
target variable $Y$ is among 
the most important goals in statistical data analysis.
However, drawing 
causal conclusions from observational data alone without intervening on the system 
is difficult. This is because
the observed  statistical dependences between $Y$ and each $X_j$ need not be due to an influence of $X_j$ on $Y$. Instead, due to Reichenbach's Principle of Common Cause \cite{Reichenbach1956}, $Y$ may also be the cause of $X_j$ or there may be a common cause $Z$ influencing both.
In many applications, time order or other prior information excludes
that $Y$ influences $X_j$. For instance, if $Y$ describes the
health condition of a patient at time $t$ and
$X_j$ some treatments at an earlier time, we `only' need to decide
to what extent the dependences between 
$X_j$  and $Y$ are due to $X_j$ influencing $Y$ and to what extent they are due to common causes ('confounders'). Here we are not interested in the reason for dependences between the variables $X_j$ themselves, we therefore merge them to a vector-valued variable $\bX$. Moreover, we restrict the attention to the case
where there is only one real-valued confounder $Z$. 
In the case of linear relations, the structural equations then read:
  \begin{eqnarray}
\bX &=& \bb Z+\bE \label{eq:xcon}\\
Y &=& \langle \ba, \bX\rangle + c Z +F\label{eq:ycon}\,.
\end{eqnarray}
where $\bE$ is a random vector with values in $\R^d$ and $Z,F$ are scalar random variables. Here, 
$\bE,Z,F$ are jointly independent, while
the components of $\bE$ may be dependent.
Here, $\ba\in \R^d$ is the vector of structure coefficients 
determining the influence of the $d$-dimensional variable $\bX$ on the scalar target variable $Y$. Likewise, $\bb\in \R^d$ is the vector determining the influence of $Z$ on $\bX$ 
and the scalar
 $c\in \R$ is the structure coefficient determining the influence of $Z$ on  $Y$.
By rescaling $\bb$ and $c$, we may assume  $Z$ to have unit variance without loss of generality.
The corresponding DAG is shown in Figure~\ref{fig:generaldag}. 
If all variables are centered Gaussian, the remaining model parameters are the vectors $\ba,\bb$, the covariance matrix $\Sigma_{\bE\bE}$ and the scalars $c,\sigma_F$, where $c$ describes the strength of the influence of $Z$ on $Y$ and $\sigma_F$ the standard deviation of $F$.  
Since this paper will be based on second-order statistics alone,
we will treat these parameters as the only relevant ones. 
The
following special cases can be obtained by appropriate choices of these parameters:

\vspace{0.2cm}
\noindent
{\bf Purely causal:}
The case $\bX\to Y$ with no confounding can easily be obtained by setting $\bb=0$ or $c=0$. In the first case, that is $c\neq 0$,  $Z$ takes the role of  an additional independent noise term (apart from $F$),
see Figure~\ref{fig:xydag}. 
\hspace{2cm}
\begin{figure}
\centering     
\subfigure[]{\label{fig:copydag}}
\resizebox{4cm}{!}{
  \begin{tikzpicture}
    \node[obs] at (4,1) (Xc) {$\bX$} ;
    \node[var] at (5,2) (Z) {$Z$} edge[->] (Xc) ;  
    \node[obs] at (6,1) (Yc) {$Y$} edge[<-] (Z) ;
    \node[anchor=center] at (4.3,1.6) {$\bb$}; 
   \node[anchor=center] at (5.65,1.6) {$=$}; 

  \end{tikzpicture} 
}
\hspace{1cm}
\subfigure[]{\label{fig:yxdag}}
\resizebox{4cm}{!}{
  \begin{tikzpicture}
    \node[obs] at (9,1) (Xa) {$\bX$};
    \node[obs] at (11,1) (Ya) {$Y$}  edge[->] (Xa);
  \end{tikzpicture}
}
\caption{(a) A special case of the purely confounded
case is obtained by setting the noise $F$ of $Y$ to $0$. Then $Y$ is an exact copy of $Z$, i.e., the cause of $\bX$. Since $P_{\bX,Y}=P_{\bX,Z}$, no
statistical method relying purely on observational
data is able to distinguish this case from
the scenario in (b), although the difference certainly matters when interventions on $Y$ are made.}
\end{figure}
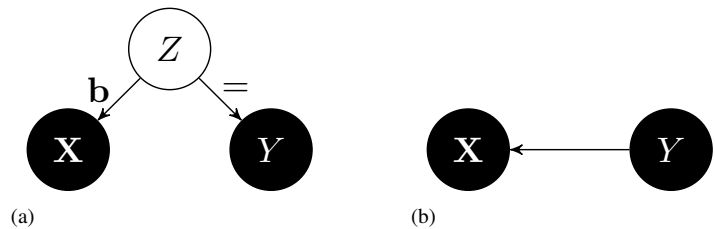

The structural equation then reads
\begin{eqnarray}
Y &=& \langle \ba, \bX\rangle + cZ +F =  \langle \ba, \bX\rangle + \tilde{F} \label{eq:xtoy}\,,
\end{eqnarray}
with $\bE,\tilde{F}$ being jointly independent.
For fixed $c$, the limit of
a deterministic influence of $\bX$ on $Y$ can be obtained by
letting at least one component of the vector $\ba$ grow to infinity. 
Then $Y$ is dominated by the term $\langle \ba, X\rangle $.

\vspace{0.2cm}
\noindent
{\bf Purely confounded:}
Setting $\ba=0$ turns the influence of $\bX$ on $Y$ off.
Then the relation between $\bX$ and $Y$ is generated by
the confounder $Z$ only, see Figure~\ref{fig:zdag}. Depending on the remaining parameters
$\Sigma_{\bE\bE}$, $\sigma_F$ and $\bb$, we obtain a scenario where 
$\bX$ provides perfect knowledge about $Z$ (when $\|\bb\|^2\to \infty$  or
$\Sigma_{\bE\bE} \to 0$) or a scenario where $Y$ provides perfect 
knowledge of $Z$ ($\sigma_F \to 0$).

\vspace{0.2cm}
\noindent
{\bf Purely anticausal:}
We have actually excluded a scenario where $Y$ is the cause of $\bX$. Nevertheless,
if $\ba=0$ and $\sigma_F=0$, we have $Y=Z$ almost surely and $Z$ is the cause of $\bX$. Hence, the scenario gets indistinguishable from an 'anticausal' scenario where $Y$ is the cause 
as in Figure~\ref{fig:yxdag}, although 
performing interventions on $Y$ would still tell us that it is not the cause.  

\vspace{0.2cm}
We now ask how to distinguish between these cases
given joint observations of $\bX$ and $Y$ that are i.i.d. drawn from $P_{\bX,Y}$. 
Conditional statistical independences,
which are usually employed for causal inference,
\cite{Pearl2000,Spirtes1993} are not able to distinguish
between the above cases since there may not be conditional independences in $P_{\bX,Y}$. 
Moreover, we assume that there are no observed causes of $\bX$ that
could act as so-called {\it instrumental} variables \cite{Bowden}
which would enable the distinction between `causal' and `confounded'. 

To see that the parameters $\ba,\bb,\se,\sigma_F$ are heavily underdetermined in linear {\it Gaussian} models, 
just note that any multivariate Gaussian $P_{\bX,Y}$
can be explained by $\bX$
being an unconfounded cause of $Y$ according to
the structural equation (\ref{eq:xtoy}) by setting
\begin{equation}\label{eq:stregress}
\hat{\ba}:=\Sigma_{\bX \bX}^{-1} \Sigma_{\bX Y}\,.
\end{equation}
The vector $\hat{\ba}$ is the vector of regression coefficients obtained by regressing $Y$ on $\bX$
without caring about the true causal structure.
Here we use the symbol $\hat{\ba}$ instead of $\ba$ to
indicate that it differs from the vector $\ba$ that  appears in the structural equation (\ref{eq:ycon})
which correctly describes the causal relation between
$\bX$ and $Y$. 
This way, we obtain a model that correctly describes the observed correlations, but not the {\it causal} relations
since the impact of interventions is not modelled correctly. -- Note that identifying $\ba$ 
is typically the main goal of causal data analysis
since $a_j$ directly describes how changing
$X_j$ changes $Y$. 
Confusing $\ba$ with $\hat{\ba}$ would be 
the common fallacy of naively attributing all
dependences to the causal influence
of $X_j$ on $Y$. To see the relation between $\hat{\ba}$ and $\ba$ 
we first find
\begin{eqnarray}
\Sigma_{\bX Y}&=&{\rm Cov}(\bX,Y)={\rm Cov}(\bE+\bb Z,\langle \ba,\bX\rangle + cZ+ F)\nonumber \\
&=& {\rm Cov}(\bE+\bb Z,\langle \ba,\bE +\bb Z\rangle + cZ+ F)\nonumber 
= (\Sigma_{\bE\bE} + \bb \bb^T)\ba + c\bb\,, \label{eq:sxywithE} 
\end{eqnarray}
where we have used the joint independence of
$\bE,Z,F$ and that $Z$ is normalized. 
Likewise,
\begin{eqnarray}
\Sigma_{\bX\bX} &=&  {\rm Cov}(\bX,\bX)={\rm Cov} (\bE+\bb Z,\bE +\bb Z)\nonumber 
=  \Sigma_{\bE\bE} +\bb \bb^T\,.
\end{eqnarray}
Due to (\ref{eq:stregress}) we thus obtain
\begin{equation}\label{eq:ahat}
\hat{\ba}=\ba +(\Sigma_{\bE\bE}+\bb \bb^T)^{-1} c \,\bb\,.
\end{equation}
Eq.~(\ref{eq:ahat}) shows that the vector $\hat{\ba}$ obtained by standard regression consists of $\ba$ (which defines the causal influence of $\bX$ on $Y$) and  a term  that is due to confounding. 

It is known that confounding can be detected in linear models with {\it non-Gaussian} variables
 \cite{HoyerLatent08}. This is because describing data generated by the model (\ref{eq:xcon}) and (\ref{eq:ycon}) by the structural equation
\[ 
Y = \langle \hat{\ba} , X\rangle  + F 
\] 
(\ref{eq:xtoy}) yields in the generic case a noise variable $F$
that is not statistically {\it independent} of $\bX$, although it is {\it uncorrelated}.
Other recent proposals to detect confounding
using information beyond conditional statistical dependences rely on different model assumptions.
Ref.~\cite{UAI_CAN}, for instance assumes non-linear relations with additive noise, while Ref.~\cite{UAI_simplex}
assumes a discrete confounder attaining a few values only. 

Here we propose a method for distinguishing the purely causal from the
confounded case that only relies 
on second order statistics and thus
does not rely on
non-Gaussianity of noise variables and higher-order
statistical independence tests like \cite{UAI_CAN}, for instance. 
The paper is structured as follows. Section~\ref{sec:newassumptions} describes the 
idea of the underlying principle, defines it formally in terms of a spectral measure and justifies it by a toy model where parameters are randomly generated.
Section~\ref{sec:defstrength} defines the strength of confounding, which is the crucial target quantity to be estimated. Section~\ref{sec:method} describes 
the method to estimate the strength and justifies it by intuitive arguments
first and by theoretical results which are rigorously shown in Section~\ref{sec:asym}.

\section{Detecting confounders by the principle of generic orientation} 

\label{sec:newassumptions}

\subsection{Intuitive idea and background}

The idea of our method is based on 
 the recently stated  {\it Principle of Independent Conditionals}  \cite{Algorithmic,LemeireJ2012} in the context of causal inference.   
To introduce it, let $G$ be a directed acyclic graph (DAG)
 formalizing the
hypothetical causal relations among the random 
variables $Z_1,\dots,Z_n$. The set of distributions compatible with
this causal structure is given by 
\[
P(Z_1,\dots,Z_n)=\prod_{j=1}^n P(Z_j|PA_j),
\]
where each $P(Z_j|PA_j)$ denotes the conditional distribution of $Z_j$, given
its parents \cite{Pearl2000}. 
Informally speaking, the Principle of Independent Conditionals
states that, usually, each $P(Z_j|PA_j)$ describes an independent mechanism of nature and therefore these objects  are `independent' and contain `no information' about each other. 
\cite{Algorithmic,LemeireJ2012} formalized `no information' by postulating that the description length of one $P(Z_j|PA_j)$ does not get shorter when
the description of the other $P(Z_i|PA_i)$ for $i\neq j$ are given. Here,
description length is defined via Kolmogorov complexity, which is, unfortunately, uncomputable \cite{Vitanyi08}. To deal with this caveat, 
one can either approximate Kolmogorov complexity, or, as shown 
in \cite{Steudel_additive_noise}, indirectly use the principle as a justification for
new inference methods rather than as an inference method itself. 

However, there are also other options to give a definite meaning to the term `independence'. To see this, consider some parametric model where 
each $P(Z_j|PA_j)$ is taken from a set of possible conditionals
$P^{\theta_j}(Z_j|PA_j)$ where $\theta_j$ is taken from some parameter space
$\Theta_j$. Assume that, for a given distribution
$P(Z_1,\dots,Z_n)$, the parameters $\theta_1,\dots,\theta_n$ are related by
an equation (i.e., one is the function of the others) that is not satisfied by
generic $n$-tuples. One can then consider this as a hint that 
the mechanisms correponding to the $P(Z_j|PA_j)$ have not been generated independently and become skeptical about the causal hypothesis\footnote{See   
\cite{LemeireJ2012}, Theorem~3, for a detailed discussion of the conditions
under which one should trust this argument.}. 
This philosophical argument is also the basis for Causal Faithfulness
\cite{Spirtes1993}, that is, the principle of rejecting a causal DAG for which
the joint distribution satisfies conditional independences that do not hold for {\it generic} vectors $(\theta_1,\dots,\theta_n)$ because it requires
the vector to lie in a lower dimensional manifold. 
In its informal version, the  Principle of Independent Conditionals generalizes this idea by excluding
also other
`non-generic' relations between parameter vectors $\theta_j$.  


We now discuss how to give a meaning to the phrase `non-generic' relation 
for our special scenario where the causal hypothesis reads
$\bX\to Y$ (without confounding and within a linear model). 
Recalling that we consider $\sx$ 
the crucial model parameter for $P_\bX$ and the regression vector $\ba$ for $P_{Y|\bX}$,
we therefore postulate that $\ba$
lies in a `generic' orientation relative to  $\Sigma_{\bX \bX}$ in a sense
to be described in Subsection~\ref{subsec:defgeneric}.
To approach this idea first by intuition, note, for instance,
that it is unlikely that $\ba$ is close to being aligned with the 
eigenvector of $\Sigma_{\bX\bX}$ corresponding to the largest
eigenvalue (i.e., the {\it first} principal component), given that $\ba$ has been chosen 'without knowing' $\Sigma_{\bX\bX}$.
Likewise, it is unlikely that it is 
approximately aligned with the {\it last}
principal component. 

For the more general DAG shown of Figure~\ref{fig:generaldag}
we again assume that $\ba$ has generic orientation with respect to the eigenspaces of $\Sigma_{\bX\bX}$ and, in addition, that
$\bb$ has generic orientation with respect to the
eigenspaces of $\Sigma_{\bE\bE}$.   

To provide a first intuition about why 
the orientation of the resulting vector $\hat{\ba}$ of regression coefficients is no longer generic relative to $\sx$
as a result of confounding, we show two somehow opposite extreme cases where
$\hat{\ba}$ gets aligned with the first and the last principal component of $\bX$, respectively. 
To this end, we
consider the
purely confounded case where $\ba=0$ and thus $\hat{\ba}=c \,\Sigma_{\bX\bX}^{-1} \bb$. 

\vspace{0.2cm}
\noindent
{\bf $\hat{\ba}$ aligned with the first eigenvector of $\sx$:} Let
$\se$ be the identity matrix $I$. Since $\se$ is then rotation invariant,
$\bb$ has certainly `generic orientation' relative to $\se$, according to any reasonable sense of `generic orientation'. 
Then $\bb$ does not have generic orientation relative to $\sx=I+\bb \bb^T$, because it is the unique eigenvector of the latter with maximal eigenvalue.
In other words, $\bb$ is aligned with the first principal component of $\bX$. 
Then, $\hat{\ba}$ is also aligned with the same principal component 
since it is a multiple of $\bb$ due to
$\hat{\ba}=\sx^{-1} \bb =(I+\bb \bb^T)^{-1} \bb$. 
Note that one also gets close to this scenario when the spectral gaps of $\se$ are small\footnote{In this limit,
the first eigenvector of $(\se +\bb \bb^T)$ is almost aligned with $\bb$. On the other hand, the vector $(\se +\bb \bb^T)^{-1} \bb$ is
almost aligned with $\bb$ because, as will be shown later,
it is a multiple of $\se^{-1} \bb$ which is almost a multiple
of $\bb$ whenever the spectral gaps of $\se^{-1}$ are negligible
compared to the spectral values of $\se^{-1}$.}  compared to the norm
of $\bb \bb^T$ and to the eigenvalues of $\se$.

\vspace{0.2cm}
\noindent
{\bf $\hat{\ba}$ close to the last eigenvector
of $\sx$:} Let the spectral gaps between adjacent eigenvalues of $\se$ be much larger than the norm of $\bb \bb^T$. Then adding
$\bb \bb^T$ changes the eigenspaces only slightly \cite{Kato}. Hence, if $\bb$ has a generic orientation relative to $\se$, it is  still generic relative to 
$\sx$. Multiplying 
$\bb$ with $\sx^{-1}$ then generates a vector that has stronger coefficients in the small eigenvalues of $\sx$. If the smallest eigenvalue of $\sx$ is much smaller than the others, the orientation of $\hat{\ba}$ gets arbitrarily 
close to the smallest eigenvector. 

\vspace{0.2cm}
For general $\se$, where the gaps between the eigenvalues are neither tiny nor huge compared to the norm of $\bb \bb^T$, the orientation of $\hat{\ba}$ changes in a more sophisticated way that heavily depends on the structure of the spectrum of $\se$.
This will be analyzed in Section~\ref{sec:method}.


\subsection{Defining `generic orientation' via the induced spectral measure}

\label{subsec:defgeneric}

We start with some notation and terminology and formally introduce two measures which have 
quite simple intuitive meanings. 
For $d\times d$ matrices, we introduce the renormalized trace\footnote{Note that $\tau$ is known as `tracial state' in the context of functional analysis \cite{Mu90}.} 
\[
\tau(A):=\frac{1}{d}{\rm tr}(A)\,.
\]
For notational convenience,
 we will assume that the spectra of all matrices are  non-degenerate throughout the paper, i.e., all eigenvalues are different. Every symmetric matrix $A$
thus admits a unique decomposition
\begin{equation}\label{eq:spectraldec}
A=\sum_{j=1}^d \lambda_j \phi_j \phi_j^T\,,
\end{equation}
where $\lambda_1> \cdots > \lambda_d$ and $\phi_1,\dots,\phi_d$ denote the corresponding eigenvectors of unit length. Every $A$ uniquely defines a measure
on $\R$, namely the distribution of eigenvalues, formally
given as follows:
\begin{Definition}[tracial spectral measure]\label{def:tracialmeasure}
Let $A$ be a real symmetric matrix with non-degenerate spectrum. 
Then the tracial spectral measure $\mu_{A,\tau}$ of $A$ is the discrete measure on $\R$ given by the uniform
distribution over its eigenvalues $\lambda_1,\dots\lambda_d$, i.e.,
\[
\mu_{A,\tau}:=\frac{1}{d}\sum_{j=1}^d \delta_{\lambda_j}\,,
\]
where $\delta_s$ denotes the point measure on $s$
for some $s\in \R$.
\end{Definition} 
By elementary spectral theory of symmetric matrices \cite{ReedSimon80},
we have:
\begin{Lemma}[expectation for tracial spectral measure]\label{lem:expectationtracial}
The expectation of any function $f:\{\lambda_1,\dots,\lambda_d\}\to \R$ with respect to the tracial measure is given as follows:
\[
\int f(w) d\mu_{A,\tau} (w)=\tau (f(A))\,.
\]
\end{Lemma}
While the spectral measure is a property of
a matrix alone, the following measure describes
the relation between a matrix and a vector:
\begin{Definition}[vector-induced spectral measure]\label{def:spectralmeasurevector}
Let $A$ be a symmetric matrix and $\lambda_j,\phi_j$ be defined by (\ref{eq:spectraldec}).
For arbitrary $\psi\in \R^d$,
 the (unnormalized) spectral measure induced by $\psi$ on $A$, denoted by $\mu_{A,\phi}$, is given by
\[
\mu_{A,\psi} (S)= \sum_{j \hbox{ with } \lambda_j \in S} \langle \psi, \phi_j\rangle^2\,,
\]
for any measurable set $S\subset \R$.
\end{Definition}
For each set of eigenvalues of $A$, the measure describes the squared length of the component
of $\phi$ that lies in the respective eigenspace of $A$.
Accordingly, we have the following normalization condition:
\begin{equation}\label{eq:normalization}
\mu_{A,\psi}(\R)= \|\psi\|^2 . 
\end{equation}
In analogy to Lemma~\ref{lem:expectationtracial} we obtain:
\begin{Lemma}[expectations for vector-induced spectral measure]\label{lem:expecationspectral}
The expectation of any function $f$ on the spectrum of $A$ with respect to 
$\mu_{A,\psi}$ 
 is given as follows:
\[
\int f(s) d\mu_{A,\psi} (s)=\langle \psi, f(A) \psi\rangle\,.
\]
\end{Lemma}
To deal with the above measures in numerical computations, 
each measure will be represented by two vectors:
first, one vector $\lambda:=(\lambda_1,\dots,\lambda_d)$ listing its support (with the convention $\lambda_1 \geq \cdots \geq \lambda_d$) and second, the vector
$w:=(w_1,\dots,w_d)$ listing the corresponding weights.
For the tracial spectral measures, $\lambda$ is just the list of eigenvalues
 and $w=(1/d,\dots,1/d)$ is just the
uniform distribution on these $d$ points.
For spectral measures induced by a vector, 
$\lambda$ is still the list of eigenvalues but
now $w$ describes the squared coefficients of the vector
with respect to the eigenvector decomposition. 

To understand our algorithm described later, it is helpful to note that using the eigenvectors $\phi_j$ of a matrix $A$ one can easily construct a vector $\psi$ that induces the tracial measure:
\[
\psi:=\frac{1}{\sqrt{d}}\sum_{j=1}^d \phi_j\,.
\]
Then we have
\begin{equation}\label{eq:inducing_tracial}
\mu_{A,\psi}=\mu_{A,\tau}\,.
\end{equation}

We are now in a position to formulate the postulate upon which
our detection of confounding is based on. The reader may feel uncomfortable
about the fact that it contains $\approx$-signs. They occur because
our probabilistic model of choosing random vectors $\bb$ independently of $\se$ and $\ba$ independently
of $\sx$ yields approximate equalities that are satisfied with high probability. 
Later this will be made mathematically precise by asymptotic statements
for the  limit of $d\to \infty$ in Section~\ref{sec:asym}. We have avoided to start with the precise statements for two reasons: first, they
require functional analysis that some reader may want to skip. Second, 
the method is applied to finite dimensional data anyway and precise statements
for finite dimensions
like 'the equality holds up to and error of $\epsilon$ with probability
at least\dots' seem even harder to get than asymptotic statements. 
\begin{Postulate}[generic orientation of vectors]\label{post:genorient}
If (\ref{eq:xcon}) and (\ref{eq:ycon}) are structural equations corresponding to the causal DAG in Figure~\ref{fig:generaldag}, and $d$ is large, then:

\vspace{0.2cm}
\noindent
(I) The vector $\ba$ has generic orientation relative to $\Sigma_{\bX\bX}$ in the sense that
\begin{equation}\label{eq:ageneric}
\mu_{\sx,\ba} \approx \mu_{\sx,\tau} \|\ba\|^2\,.
\end{equation}

\vspace{0.2cm}
\noindent
(II) The vector $\bb$ has generic orientation relative to
$\Sigma_{\bE\bE}$ in the sense that
\begin{equation}\label{eq:bgeneric}
\mu_{\se,\bb} \approx \mu_{\se,\tau} \|\bb\|^2\,.
\end{equation}

\vspace{0.2cm}
\noindent
(III) The  vector $\ba$  is generic relative to $\bb, \se$ in the sense that 
\begin{equation} \label{eq:genericvecvec}
\mu_{\sx,\ba+c\,\sx^{-1}\bb} \approx \mu_{\sx,\ba} +
\mu_{\sx,c\,\sx^{-1} \bb}\,.
\end{equation}
\end{Postulate}
We have not explained yet why the above three conditions can be seen as
being implied by some kind of 'genericity' assumption.
The final justification will be given in Section~\ref{sec:asym} by the proof
of Theorem~\ref{thm:justi} stated in Subsection~\ref{subsec:justi}, but we provide
some rough arguments now. 

Intuitively speaking,  (\ref{eq:ageneric})
states that decomposing $\ba$ into eigenvectors of
$\sx$  yields weights that are close to being uniformly spread over the spectrum. Likewise,  (\ref{eq:bgeneric}) states that the weights of
$\bb$ are close to being uniformly spread over the spectrum of $\se$.
(\ref{eq:genericvecvec}) states that the spectral measure induced by $\hat{\ba}=\ba+\sx^{-1} \bb$
decomposes approximately into the part induced by
the causal vector $\ba$ and the confounding vector $\sx^{-1} \bb$. This insight will be crucial for both algorithms described in the present paper.
To see why (\ref{eq:genericvecvec}) happens to be true whenever $\ba$ is generic relative to $(\bb,\se)$, note that 
for any measureable function $g$ we have
\begin{eqnarray*}
\int g(s) d\mu_{\bX,\hat{\ba}}(s) &=& \langle \hat{\ba},g(\sx) \hat{\ba}\rangle\\  
&=& \langle \ba , g(\sx) \ba \rangle + \langle \sx^{-1} \bb , g(\sx) \sx^{-1} \bb \rangle 
+
 2\langle \sx^{-1} \bb,  g(\sx)  \ba \rangle \\ 
&\approx & \langle \ba , g(\sx) \ba \rangle + \langle \sx^{-1} \bb , g(\sx) \sx^{-1} \bb \rangle \\
&=& \int g(s) d\mu_{\bX,\ba }(s) +\int g(s) d\mu_{\bX,\sx^{-1} \bb}(s)\,,
\end{eqnarray*}
because $\langle \ba, g(\sx) \sx^{-1} \bb\rangle \approx 0$ if $\ba$ is in generic orientation relative to the vector
$g(\sx) \sx^{-1} \bb$.

Note that (\ref{eq:ageneric}), (\ref{eq:bgeneric}), and
(\ref{eq:genericvecvec}) 
only hold if the $\approx$-sign is interpreted in a sufficiently loose sense.
This will later be made precise within a model 
where the differences between both sides of the above equations converge
{\it weakly} to zero. They are {\it not} close, for instance, 
with respect to total variation distance.
Hence, the measures are similar
in the same sense as two empirical distributions
with large sample size 
 are similar when they
are independently sampled from the same distribution.

\vspace{0.2cm}
\noindent
{\bf Relation to the Trace Condition:} We now describe the relation of the above ideas to those underlying the
so-called {\it Trace Method} \cite{tracemethod,frW_UAI}, which is, to the best of our knowledge, the
work in the literature that is closest to the present one.
Let $\bX$ and $\bY$ be vector-valued variables with values in $\R^d$ and $\R^m$, respectively. Assume $\bX$ influences $\bY$ via the linear model
\[
\bY = A\bX +\bF\,,
\]
where $A$ is an $m\times d$ matrix and $\bF$ a noise variable of dimension $m$. 
Then $A$ and $\sx$ satisfy the trace condition
\begin{equation}\label{eq:tracecondition}
\frac{1}{m}{\rm tr} (A \sx A^T) \approx \frac{1}{d} {\rm tr} (\sx) \frac{1}{m} {\rm tr}(A A^T)\,. 
\end{equation}
For $m=1$, we can replace the $1\times d$ matrix $A$ with the vector $\ba$
and $A\bX$ with the inner product $\langle \ba, \bX\rangle$. Then
(\ref{eq:tracecondition}) turns into
\begin{equation}\label{eq:firstmoments}
\langle \ba, \sx \ba \rangle \approx \frac{1}{d} {\rm tr} (\sx) \langle \ba, \ba\rangle\,. 
\end{equation}
In terms of the spectral measures, (\ref{eq:firstmoments})  reads
\[
\int s \,d\mu_{\sx,\ba} (s) \approx \int s\, d\mu_{\sx,\tau}(s) \|\ba\|^2\,.
\]
Hence, (\ref{eq:tracecondition}) postulates 
that the {\it first moments} of two measures on the left and the right
of (\ref{eq:ageneric}) coincide almost, while our method also accounts for
{\it higher order} 
moments which the Trace Condition ignores.
As already sketched in \cite{frW_UAI}, the Trace Condition (\ref{eq:tracecondition}) is closely related to 
the concept of {\it free independence}
in free probability theory \cite{Voiculescu}.
In the appendix we will explain why 
(\ref{eq:ageneric}), (\ref{eq:bgeneric}), (\ref{eq:genericvecvec}) are also related to free independence in spirit, although
there is no straightforward way to apply those concepts here.

\subsection{Justifying the postulates by a generating model}

\label{subsec:justi}

We now define the following sequence of models for increasing dimension $d$ for which the approximate equalities (\ref{eq:ageneric}), (\ref{eq:bgeneric}),
and (\ref{eq:genericvecvec}) become equalities in the limit $d\to \infty$:

\vspace{0.2cm}
\noindent
{\bf Covariance matrix of the noise of $\bX$:} 
Let $(\se^d)_{d\in \N}$ be a uniformly bounded sequence of positive semi-definite $d\times d$-matrices such that their tracial spectral measures  converge weakly to
some measure $\mu^\infty$ (describing the asymptotic
distribution of eigenvalues).

\vspace{0.2cm}
\noindent
{\bf  Vector of causal structure coefficients $\ba$:}
Let $(\ba_d)_{d\in \N}$ be a sequence of  vectors 
in $\R^d$ drawn uniformly at random from a sphere of
fixed radius $r_\ba$.

\vspace{0.2cm}
\noindent
{\bf  Vector of confounding structure coefficients $\bb$:}
Let $(\bb_d)_{d\in \N}$ be a sequence of  vectors 
in $\R^d$ drawn uniformly at random (independently of $\ba_d$) from a sphere of fixed radius $r_\bb$.
Let $c$ be fixed for all $d$.

\vspace{0.2cm}
\noindent
Then $\sx^d=\se^d+\bb_d\bb_d^T$ and $\hat{\ba}_d=\ba_d+c (\sx^d)^{-1} \bb_d$ and
we have the following result that will be shown
in Section~\ref{sec:asym}:
\begin{Theorem}[justifying Postulate~\ref{post:genorient} by rotation-invariant generating model]\label{thm:justi}
For the above generating model the 
approximations
(\ref{eq:ageneric}), (\ref{eq:bgeneric}),
and (\ref{eq:genericvecvec}) are asymptotically justified in the sense that
\begin{align}
\mu_{\sx^d, \ba_d} &\rightarrow r^2_\ba \mu^\infty \quad &\hbox{ (weakly in probability) } \label{eq:asympta}\\
\mu_{\se^d, \bb_d} &\rightarrow r^2_\bb \mu^\infty
\quad &\hbox{ (weakly in probability) } \\
\mu_{\sx^d, \ba_d + c (\sx^d)^{-1} \bb_d } -
(\mu_{\sx^d, \ba_d} + \mu_{\sx^d,c(\sx^d)^{-1}\bb_d })
&\rightarrow 0 \quad &\hbox{ (weakly in probability) }
\label{eq:asymporth}
\end{align}
\end{Theorem}
The above highly symmetrical generating model may appear as a too strong assumption for practical purposes. It should therefore be noted that 
much weaker assumptions would probably also yield
the same approximate identities for high dimensions.
We therefore built our algorithm in Section~\ref{sec:asym} upon the postulates only
instead of directly using the generating model.

\section{Characterizing confounding by two parameters \label{sec:defstrength}}

\subsection{Strength of confounding}

To understand to what
extent the dependences between $\bX$ and $Y$ is due to the influence of
$\bX$ on $Y$ and to what extent it is due to confounding, we first introduce the following parameter that quantifies the relative contribution of the confounding to the covariance of $\bX$ and $Y$:
\begin{Definition}[correlative strength of confounding]
In the structural equations (\ref{eq:xcon}) and (\ref{eq:ycon}) 
the correlative strength of confounding is given by
\begin{equation}\label{eq:defcorrelativestrength}
\gamma:=
\frac{\|\Sigma_{\bX Z}\|^2}{\|\Sigma_{\bX Y}\|^2 + \|\Sigma_{\bX Z}\|^2}=
 \frac{\|c \cdot \bb\|^2}{\|\sx \ba\|^2 + \|c\cdot \bb\|^2}
\,.
\end{equation}
\end{Definition}
Note that the first formulation of $\gamma$ 
on the right hand side of \eqref{eq:defcorrelativestrength} refers to quantities that were directly observable when $Z$ would be observable. Here, we have considered $\Sigma_{\bX Z}$ 
as  {\it vector} with the $d$ entries ${\rm Cov}(X_j,Z)$.

The second definition of $\gamma$, on the other hand, gets a particularly simple meaning when (\ref{eq:genericvecvec}) in Postulate~\ref{post:genorient} holds.  Then 
\begin{equation}\label{eq:strengthorth}
\|\sx \ba\|^2+\|c \cdot \bb\|^2 \approx  \|\Sigma_{\bX Y}\|^2\,. 
\end{equation}
Hence, $\|\Sigma_{\bX Y}\|^2$ (which quantifies the covariance between $\bX$ and $Y$)
is a sum of the term quantifying the confounding and a term quantifying the causal influence of $\bX$ on $Y$.
Hence, $\gamma$ measures which fraction of the squared covariance 
is caused by confounding:
\begin{equation}\label{eq:gammaApprox}
\gamma \approx \frac{\|c\cdot \bb\|^2}{\|\Sigma_{\bX Y}\|^2}\,.
\end{equation}

We now focus on a different definition of strength of
confounding that measures how much $\hat{\ba}$ deviates from $\ba$ (relative to the sum of the squared lengths of these vectors):
\begin{Definition}[structural strength of confounding]\label{def:structuralstrength}
The structural strength of confounding is
defined by
\begin{equation}\label{eq:defstructuralstrength}
\beta:=
\frac{\|\sx^{-1} \Sigma_{\bX Z}\|^2}{\|\Sigma_{\bX \bX}^{-1} \Sigma_{\bX Y}\|^2 +\|\sx^{-1} \Sigma_{\bX Z}\|^2}=
\frac{\|c \cdot \sx^{-1} \bb\|^2}{\|\ba\|^2 +\|c \cdot \sx^{-1} \bb\|^2}  \,.
\end{equation}
\end{Definition}
Again, the denominator can be approximated via
\begin{equation}\label{eq:betaApp}
\|\ba\|^2 +\|c \cdot \sx^{-1} \bb\|^2 \approx \|\hat{\ba}\|^2\,,
\end{equation}
due to eq.~(\ref{eq:genericvecvec}) and \eqref{eq:normalization} because 
\[
\|\hat{\ba}\|^2=\mu_{\sx,\hat{\ba}}(\R) \approx \mu_{\sx,\ba}(\R) +\mu_{\sx, c\,\sx^{-1} \bb}(\R) =
\|\ba\|^2 + \|c\,\sx^{-1} \bb\|^2. 
\]

The relation between $\beta$ and $\gamma$ is quite non-trivial
 and depends on many other quantities as we will see below. Remarkably, they can differ by orders of magnitudes\footnote{This is an interesting phenomenon in high dimensions: a confounder may
generate almost no observable covariance between $\bX$ and $Y$ and still 
perturb the vector of regression coefficients significantly.}. Without claiming that $\beta$ would be the better measure\footnote{Note that quantifying causal influence in causal Bayesian networks is non-trivial and there exists no generally accepted measure \cite{causalstrength}.}, we focus on $\beta$ because 
it is more relevant for causal statements: whenever
$\beta$ is large identifying $\hat{\ba}$ with $\ba$ yields significantly wrong causal conclusions
even when $\gamma$ is small. We have introduced $\gamma$ only to show that whether
confounding is negligible or not highly depends on how it is quantified. 

Note that $\beta$ and $\gamma$ are related via
\[
\beta=\gamma \frac{\|\Sigma_{\bX Y}\|^2}{\|\hat{\ba}\|^2}\cdot \frac{\|\sx^{-1} \bb\|^2}{\|\bb\|^2} \,.
\]
Unfortunately, the factor  $\|\sx^{-1} \bb\|/\| \bb\|$ cannot be seen from observations alone since $\bb$ is not observed.
However, 
subject to Postulate~\ref{post:genorient}, we
obtain a non-linear relation between $\beta$ and $\gamma$ 
depending on the spectrum of $\se$. We describe this relation in the appendix.

\subsection{A second parameter characterizing confounding}

The contribution of $Z$ to the covariance between $\bX$ and $Y$ is given by the 
product $c \cdot \bb$. Accordingly, rescaling $\bb$ with some factor while rescaling $c$ 
with the inverse factor preserves the correlative strength of confounding. 
Although structural confounding strength is affected by rescaling $\bb$ and $c$ in a more sophisticated way, $\beta$ can also be unaffected by rescaling both $\bb$ and $c$ in an appropriate way. 
The regimes with small $\bb$ and large $c$  versus
the one with large $\bb$ and small $c$ have simple interpretations: in the first case,
the uncertainty of $\bX$ is hardly reduced by knowing $Z$, while in the second case, knowing
$Z$ reduces most of the uncertainty of $\bX$. 

To distinguish between these different regimes of confounding we introduce the second parameter $\eta$, which
measures  the explanatory power of $Z$ for $\bX$:
\begin{eqnarray*}
\eta &:=&{\rm tr} (\sx) - {\rm tr} (\Sigma_{\bX\bX |Z})
= {\rm tr} (\sx) - {\rm tr} (\se)= \|\bb\|^2 \leq \|\sx\|.
\end{eqnarray*}

For the entire Section~\ref{sec:defstrength} it is important to keep and mind that
we always  referred  to the case where
$Z$ has unit variance.


\section{Description of the method}

\label{sec:method}

\subsection{Constructing typical spectral measures for given parameter values}

\label{subsec:constrnubeta}

The main result of this section states that asymptotically, $\mu_{\sx,\hat{\ba}}$
depends only on $3$ parameters when $\sx$ is given: $\|\hat{\ba}\|^2$, $\beta$, $\eta$.
The first one is directly observable, hence we define a two-parametric family
of normalized (i.e. probability) measures
$\nu_{\beta,\eta}$ such that for large $d$ with high probability
\[
\mu_{\sx,\hat{\ba}} \approx \|\hat{\ba}\|^2 \nu_{\beta,\eta}.
\]
We first  describe the construction of $\nu_{\beta,\eta}$:

\begin{enumerate}

\item {\bf Causal part:}  this part describes the spectral measure 
that were obtained in the absense of confounding.
It is induced by $\ba$ and
$\sx$. According to \eqref{eq:ageneric}, it is approximated by the uniform distribution over the spectrum of $\sx$, i.e., the tracial measure introduced in Definition~\ref{def:tracialmeasure}.
We therefore define:
\[
\nu^{\rm causal}:=\mu_{\sx,\tau}.
\] 

\item {\bf Confounding part:} we now approximate the spectral measure induced by the vector $\sx^{-1} \bb$ and $\sx$. 
We will justify the construction later after we have described all steps.
We first define the matrix 
$M_X:= {\rm diag} (v^X_1,\dots,v^X_d)$, where $v_j^X$ are the eigenvalues of $\sx$ in decreasing order. Then we define a rank-one perturbation of
$M_X$ by 
\begin{equation}\label{eq:TdefFinite}
T:= M_X + \eta \bg \bg^T, 
\end{equation}
where $\bg$ is the vector ${\bf g}:=(1,\dots,1)^T/\sqrt{d}$. 
We then compute the spectral measure induced by the vector $T^{-1} \bg$ and
$T$ and define 
\begin{equation}\label{eq:nuconfounded}
\nu_{\eta}^{\rm confounded}:= \frac{1}{\|T^{-1} \bg\|^2} \mu_{T, T^{-1} \bg}. 
\end{equation}

\item {\bf Mixing both contributions:} Finally, $\nu_{\beta,\epsilon}$ is a convex sum of the causal and the confounded part where the mixing weight of the latter is given by the confounding strength:
\[
\nu_{\beta,\eta}:=(1-\beta) \nu^{\rm causal} +\beta \nu^{\rm confounded}_{\eta}.
\]
\end{enumerate}

We now explain Step~2 in the above construction. According to Postulate~\ref{post:genorient}, $\bb$ has generic orientation relative to $\se$ in the sense of \eqref{eq:bgeneric}. 
 With respect to its eigenbasis, $\se$
reads $M_E:={\rm diag}(w^E_1,\dots,v^E_d)$, where
$v^E_j$ are the eigenvalues of $\se$. Of course we don't know the vector $\bb$, neither do we know the coordinates of $\bb$ with respect to this basis.  
Remarkably, it turns out that knowing that $\bb$ is generic relative 
to $\se$ is enough because then we can replace $\bb$ with a vector that is
`particularly generic', namely $\bg$.  This vector 
 satisfies 
\begin{equation}\label{eq:exactgen}
\mu_{M_E,\bg} = \mu_{M_E,\tau},
\end{equation}
(see \eqref{eq:inducing_tracial})
while asymptotically the overwhelming majority of vectors satisfy \eqref{eq:exactgen} {\it approximately}. 
Therefore, an appropriate multiple of $\bg$ nicely mimics the behaviour of generic vectors. 
Accordingly, we can
approximate the spectral measure induced by
$(\se+\bb \bb^T)^{-1} \bb$ and $(\se +\bb \bb^T)$ by 
the spectral measure induced by $(M_E + \eta \bg \bg^T)^{-1}
\bg$ and 
$M_E + \eta \bg \bg^T$. 
Unfortunately, this construction would need the eigenvalues of $\se$, which cannot be computed from observing $\bX,Y$ alone. 
Asymptotically, however, the difference between the spectra of $\sx$ and
 $\se$ do not matter and we can replace $M_E$ with $M_X$. This step will be justified later using 
the fact that for large $d$  we have
\begin{equation}\label{eq:tracialmeasurescoincide}
\mu_{\sx,\tau} \approx \mu_{\se,\tau}\,,
\end{equation}
which is made more precise by the
following result:
\begin{Lemma}[tracial measures are close]\label{lem:tracialclose}
For any interval $[r,l]$ we have
\[
|\mu_{\bX,\tau}[r,l] -\mu_{\bE,\tau}[r,l]|\leq d\,.
\]
\end{Lemma}
{\bf Proof:} If $v^E_1>\cdots >v^E_d$ denote the eigenvalues of $\Sigma_{\bE\bE}$, then the eigenvalues $v^X_j$  of $\sx=\Sigma_{\bE\bE}+\bb\bb^T$ 
satisfy 
\[
v^X_{j} \in [v^E_j,v^E_{j-1}] \quad \forall
j\geq 2\,,
\]
by
Theorem~10.2 in \cite{Datta}. Hence the number of eigenvalues in a given interval can differ by $1$ at most. 
$\Box$

\vspace{0.3cm}
\noindent
We now describe the main theoretical result of this article:
\begin{Theorem}[congervence to two-parametric family]\label{thm:relgenericfamily}
Let $(\sx^d)$ be a sequence of covariance matrices
for which $\mu_{\sx^d,\tau}$ converges weakly to 
some measure $\mu^\infty$ supported by a compact interval in $\R_0^+$.
Assume, moreover, we are given sequences of model parameters 
 $(\ba_d)$  and $(\bb_d)$ with $\|\ba_d\|=r_\ba$  and
$\|\bb_d\|=r_\bb$ and fixed $c$, such that \eqref{eq:asympta}-\eqref{eq:asymporth} hold (recall
$\se^d=\sx^d - \bb_d \bb_d^T$).
Then $\nu_{\beta,\eta}$ approximates $\mu_{\sx, \hat{\ba}}$ up to normalization in the sense
that 
\[
\frac{1}{ \|\hat{\ba}_d\|^2 }\mu_{\sx^d,\hat{\ba}_d} - \nu^d_{\beta,\eta} \to 0 \quad \hbox{ (weakly in probability) },
\]
where $\eta:=r_\bb^2$ and $\beta:= \frac{c^2 M[R[\eta \mu^\infty]](\R)}{c^2 M[R[\eta \mu^\infty]](\R) + r_\ba^2}$.
\end{Theorem}
Hence, the theorem states that whenever
Postulate~\ref{post:genorient} holds with sufficient
accuracy and for sufficiently high dimension,
then $\nu_{\beta,\epsilon}$ is a good approximation for
the induced spectral measure.
We will prove Theorem~\ref{thm:relgenericfamily} in Section~\ref{sec:asym}.

Apart from this weak convergence result we also know that
the measures $\mu_{\sx^d,\hat{\ba}_d}$ and  $\nu^d_{\beta,\eta}$
have precisely the same support for any $d$ because, by construction, 
 $\nu^d_{\beta,\eta}$ is also supported by the spectrum of $\sx$. 
This enables to conveniently represent both measures by 
vectors whose entries describe the weight of the corresponding eigenvalue.

\subsection{Description of the algorithm}

To estimate the confounding parameters we just take the element in the family
$(\nu_{\beta,\eta})_{\beta \in [0,1],\eta\in [0,v_1^X]}$ that is  closest to $\mu_{\sx,\hat{\ba}}$, but we have to choose an appropriate distance measure. 
Since Theorem~\ref{thm:relgenericfamily} only guarantees weak convergence,
$l_1$ or $l_2$ distances between the weight vectors would be inappropriate.  
Instead, we have decided to smoothen the measures
by a Gaussian kernel and then compare the $l_1$ distance.
As kernel bandwidth we have worked with $\sigma:=0.2 (v^X_1 -v^X_d)$.
Accordingly, we define a distance between two weight vectors $w$ and $w'$ by 
\begin{equation}\label{eq:Ddef}
D(w,w'):=\|K (w-w')\|_1,
\end{equation}
where $K$ denotes the kernel smoothing matrix with entries
\[
K(i,j):=e^{-\frac{(v_i^X-v_j^X)^2}{2\sigma^2}}
\]

\vspace{0.2cm}
Based on these findings,
we describe how to estimate $\beta$
in Algorithm~\ref{alg:eststrength}.
\begin{algorithm}[!htb]
\caption{Estimating the strength of confounding}
\label{alg:eststrength}
\begin{algorithmic}[1]
  \STATE {\bfseries Input:} I.i.d. samples from
$P(\bX,Y)$.  
   \vspace{0.0cm}
   \STATE  
Compute the empirical covariance matrices
$\sx$ and $\Sigma_{\bX Y}$
   \STATE Compute the regression vector 
$
\hat{\ba}:=\sx^{-1} \Sigma_{\bX Y}
$
\vspace{0.2cm}
   \STATE PHASE 1: Compute the spectral measure $\mu_{\sx,\hat{\ba}}$
   \STATE Compute eigenvalues $v^X_1 > \cdots >v^X_d$ and the corresponding eigenvectors $\phi_1,\dots,\phi_d$ of $\sx$
   \STATE Compute the weights $w'_j=\langle \hat{\ba},\phi_j \rangle^2$
     and then the normalized weights $w_j:=w_j'/\sum_j w_j'$.  
\vspace{0.2cm} 
  \STATE PHASE 2: find the parameter values $\hat{\beta},\hat{\eta}$ 
that minimize the distance $D(w,w^{\beta,\eta})$ with $D$ defined by \eqref{eq:Ddef}, where 
$w^{\beta,\eta}$ denotes the weight vector of the measure $\nu_{\beta,\eta}$.
   \STATE {\bfseries Output:} Estimated confounding strength $\hat{\beta}$
\end{algorithmic}
\end{algorithm}

Since the pseudocode does not describe how to compute the weight vector $w^{\beta,\eta}$ 
 we provide this missing detail now. 
First compute the matrix $T$ as defined by \eqref{eq:TdefFinite}
and compute its eigenvectors $\psi_1,\dots,\psi_d$.  Then we compute
the vector $v:=T^{-1} \bg/\|T^{-1} \bg\|$. The squared coefficients of $v$ with respect to the
eigenvector  basis $\phi_1,\dots,\phi_d$ describe the weights of $\nu^{\rm confounded}_\eta$, see \eqref{eq:nuconfounded}.
To obtain the weights of $\nu_{\beta,\eta}$ we need to add the  contribution of $\nu^{\rm causal}$ and
finally obtain the weights
\[
w_j^{\beta,\eta}:= \frac{1}{d}(1-\beta) + \beta \langle v, \phi_j\rangle^2.
\]

\subsection{Remark on normalization}

So far we have ignored the case where the variables $X_1,\dots,X_d$ 
refer to quantities that are measured in different units. 
If, for instance,  $X_1$ denotes the temperature and $X_2$ the traffic density
(which both influence the $NO_x$ concentration in the air), the relative scale of their numeric values depend on the units one choses. 
Another related issue is that all $X_j$ refer to the same unit, but the variance of one of the variables is overwhelmingly larger than the variance
of the others, which results in a covariance matrix whose rank is basically one.
A pragmatic and straightforward solution for both issues is to normalize all variables $X_j$ as preprocessing step. Actually, this is obviously in conflict with the justification of the method because normalization jointly changes $\sx$ and $\ba$, which spoils the idea of `independence'. 
In our simulation studies, however, the results turned out to be surprisingly robust with respect to normalizing all $X_j$. Here, robustness is only meant in the sense that the performance over a large number of runs looked almost the same. 
For every single experiment, however, the estimated values $\hat{\beta}$
can significantly differ by the amount of uncertainty that is inherent to our method anyway. Due to the lack of theoretical justification, we recommend 
to avoid normalization if possible and remain skeptical about the results  
with normalized data.

\section{Proofs of asymptotic statements} 

\label{sec:asym}

\subsection{Proof of Theorem~\ref{thm:justi}}

To show that the difference between the left and the right hand sides of
(\ref{eq:ageneric}) and 
(\ref{eq:bgeneric}), respectively, converge to zero weakly in probability, it is sufficient to show the following result:
\begin{Lemma}\label{lem:aasympt}
Let $(A_d)_{d\in \N}$ with $\|A_d\| \leq a$ be a sequence of symmetric matrices whose spectral measure converges weakly to some $\mu^{\infty}$, i.e.,
\begin{equation}\label{eq:convtracial}
\mu_{A_d,\tau} \to \mu^\infty\,.
\end{equation}
 Let $(\bc_d)_{d\in \N}$ with $\bc_d\in \R^d$ be randomly drawn from the sphere of radius $r$. Then
\[
\mu_{A_d,\bc_d} \to r^2 \mu^{\infty}\,,
\]
weakly 
in probability.
\end{Lemma}
{\bf Proof:}
It is sufficient to show the statement for $r=1$ because the measure obviously scales quadratically in $r$. 
Since the support of $\mu_{A_d,\bc_d}$ is contained in the compact interval $[-a,a]$, 
it is sufficient to show convergence of all moments, i.e., that for every $k\in \N$ 
\[
\int s^k d\mu_{A_d,\bc_d}(s) \to \int s^k \mu^\infty(s)\,,
\]
in probability.
To this end, we drop most indices $d$ and consider fixed dimension $d$. To generate a random
unit vector $\bc$, we first take $d$ independent Gaussian random variables $C_j\sim N(0,1/\sqrt{d})$ and define the $j$th coefficient of $\bc$  by 
$C_j/\sum_{i=1}^d C^2_i$. 
Let $A={\rm diag}(\lambda_1,\dots,\lambda_d)$ 
without loss of generality. Then the $k$th moment reads: 
\begin{equation}\label{eq:momentsasquotient}
\int s^k d\mu_{A,\bc}(s) =\frac{\sum_{j=1}^d \lambda_j^k C^2_j  }{\sum_{j=1}^d C^2_j}=:\frac{\Lambda_d}{\Gamma_d}\,.
\end{equation}
One easily checks 
\[
\Exp[\Lambda_d]= \sum_{j=1}^d \frac{\lambda^k_j}{d} =\int s^k d\mu_{A_d,\tau} (s)\,.
\]
Moreover, since all $C_j$ are independent and because squared standard Gaussians have variance $2$, we have
\[
{\rm Var}[\Lambda_d]= \sum_{j=1}^d \lambda_j^{2k}  {\rm Var} [C_j^2]=   \frac{2}{d^2} 
\sum_{j=1}^d \lambda_j^{2k}\leq \frac{2}{d} a^{2k}\,,
\]
where we used $\lambda_j \leq a$.
By Chebyshev's inequality, the probability
for large deviations from the mean can be bounded
by 
\begin{equation}\label{eq:cheby}
{\rm Pr}\left\{ |\Lambda_d -\int s^k d\mu_{A,\tau}(s)| \geq \epsilon \right\} \leq \frac{2}{d \epsilon^2 } a^{2k} \,.
\end{equation}
Then we get
\begin{eqnarray}
&&\int s^k d\mu_{A_d,\bc_d}(s) -\int s^k d\mu^\infty (s) \nonumber 
= \frac{\Lambda_d}{\Gamma_d} -\int s^k d\mu^\infty (s) \nonumber \\
&=& \frac{\Lambda_d - \int s^k d\mu_{A_d,\tau}(s) }{\Gamma_d} \label{eq:firstquotient} +
\left(\frac{\int s^k d\mu_{A_d,\tau} (s)}{\Gamma_d} -\int s^k d\mu^\infty(s)\right)\label{eq:tracialdiff}\,.  
\end{eqnarray}
We have $\Gamma_d\to 1$ almost surely by the strong law of large numbers.  
Due to (\ref{eq:cheby}), the term (\ref{eq:firstquotient}) converges to
zero in probability.  
Thus, expression (\ref{eq:tracialdiff}) converges
to zero in probability due to the assumption $\int s^k d\mu_{A_d,\tau} (s)\to
\int s^k d\mu^\infty(s)$. 
$\Box$

\vspace{0.2cm}
\noindent
To show that the difference between the left and the right hand side of (\ref{eq:genericvecvec}) converges weakly to zero in probability, we 
recall that it is sufficient to show that expectations of bounded continuous
functions converge. 
For any measurable function $g:R\rightarrow \R$  the difference of expectations reads:
\[
\int g \,d \mu_{\sx, \ba + \sx^{-1} \bb} - \int   g\, d\mu_{\sx, \ba} -
\int g \,d\mu_{\sx, \sx^{-1} \bb} = 2 \langle \ba , g(\sx) \sx^{-1} \bb \rangle.  
\] 
Hence, we
only have to show
that
\begin{equation}\label{eq:orthInner}
\langle \ba, g(\sx) \sx^{-1}\bb\rangle \to 0
\end{equation}
in probability. Note that this already follows
from the fact that $\ba$ is chosen independently
from the vector on the right hand side in the inner product \eqref{eq:orthInner}, due to the
following elementary result, which is probably known in the literature:
\begin{Lemma}[asymptotic orthogonality]\label{lem:simpleorth}
Let ${\bf v}_d\in \R^d$ be a sequence of vectors. Let $\bc_d\in \R^d$ be 
 drawn uniformly at random from the unit sphere. Then, 
\[
\langle {\bf v}_d,\bc_d \rangle^2 \to 0\,, 
 \,.
\]
almost surely. 
\end{Lemma}
{\bf Proof:}
Without loss of generality, assume
 ${\bf v}=(v,0,\dots,0)^T$ with $v\in \R$.
Generate the entries $c_j$ of $\bc$ by first taking
independent standard Gaussians $C_j$ and renormalizing afterwards.
Then 
\[
\lim_{d\to \infty} \langle {\bf v}_d,\bc_d \rangle^2 = 
 \lim_{d\to \infty} \frac{1}{d} \frac{ C_1^2}{\frac{1}{d} \sum_{j=1}^d C^2_j} = 0,
\]
because $\frac{1}{d} \sum_{j=1}^d C^2_j$ converges to $\Exp[C_j^2]=1$ almost surely due to the
law of large numbers.
$\Box$

\subsection{Proof of Theorem~\ref{thm:relgenericfamily}}

We first need some definitions and tools. The following one
generalizes Definition~\ref{def:spectralmeasurevector}
to infinite-dimensional Hilbert spaces (see \cite{ReedSimon80} for spectral
theory of self-adjoint operators):
\begin{Definition}[vector-induced spectral measure]
Let $\cH$ be a Hilbert space and
$A: \cH \rightarrow \cH$ a self-adjoint operator with
spectral decomposition
$A=\int \lambda dE_\lambda$, where $(E_\lambda)_{\lambda \in \R}$ denotes the
spectral family of $A$ (that is, $E_\lambda$ projects
onto the spectral subspace corresponding to all
spectral values not larger than $\lambda$).
 For any $\psi \in \cH$, let
$\mu_{A,\psi}$ be defined by
\[
\int f d\mu_{A,\psi}= \int f d \langle \psi , E_\lambda \psi\rangle,
\]
for all measurable functions $f:\R\rightarrow \R$.
\end{Definition}
We then
define a map on the space of measures
that will be a convenient tool for the proof:
\begin{Definition}[rank one perturbation for general measures]\label{def:rankonepertgeneral}
Let $\nu$ be a (not necessarily normalized)
finite measure on $\R$. Let $L^2(\nu,\R)$ denote the
Hilbert space of real square integrable functions on $\R$.
Define an operator $B_\nu $ on $L^2(\nu,\R)$ by
\begin{equation}\label{eq:Tdefgeneral}
B_\nu :={\rm id} + {\bf 1}\langle {\bf 1}, . \rangle\,,
\end{equation}
where ${\rm id}$ denotes identical map $s\mapsto s$ and ${\bf 1}$ the constant function $1$. Then 
\begin{eqnarray*}
R (\nu)&:=&\mu_{B_\nu, {\bf 1}}\,.
\end{eqnarray*}
\end{Definition}
The name `rank one perturbation' is justified because
$R$ describes how the spectral measure
induced by an operator $A$ and a vector $\psi$
changes by replacing $A$ with its rank-one perturbation
$A +\psi \psi^T$. To see this, let
$\nu = \mu_{A,\psi}$. Assume, without loss of generality, that $\psi$ is a cyclic vector for $A$, i.e., that the span of
$\{A^k \psi\}_{k\in \N}$ is dense in $\cH$ (otherwise we restrict $A$ to the completion of this span). By standard spectral theory
of operators \cite{ReedSimon80}, there is a unitary map $U: \cH \rightarrow
L^2(\R,\mu_{A,\psi})$ 
`diagonalizing' $A$ in the sense that
$A=U^* {\rm id} U$ and
$U \psi = {\bf 1}$. Therefore,
\[
\mu_{{\rm id}+ {\bf 1}{\bf 1}^T ,{\bf 1}}
=\mu_{A +\psi \psi^T, \psi}.   
\]

We do not have a more explicit description of $R$, but the relation between
the Cauchy transforms of $\nu$ and $R(\nu)$ is remarkably simple. To describe the relation, we first introduce Cauchy transforms
 \cite{CauchyTransform}:
\begin{Definition}[Cauchy transform]
Let $\nu$ be a not necessarily normalized measure.
Then the Cauchy transform of $\nu$ is defined\footnote{Note that some authors define the Cauchy transform as the negative of the below definition.} as the complex-valued function from $\C^+$ (that is, the set of complex numbers with positive imaginary part) to $\C$ 
given by
\[
F_\nu (z):=\int (z-t)^{-1} d\nu (t)\,. 
\]
\end{Definition}
Then we find:
\begin{Lemma}[spectral measure for rank-one perturbation]\label{lem:cauchywithbwithoutb}
Let $A$ be a self-adjoint operator on some Hilbert space $\cH$ and let $\bc\in \cH$ be some vector. Define
the rank one perturbation $A_\bc :=A+ \bc\langle \bc,.\rangle$. Set $\nu:=\mu_{A_\bc,c}$. Then
\begin{equation} \label{eq:AronsKrein}
F_{\nu}= \frac{F_{\mu_{A,\bc}}}{1- F_{\mu_{A,\bc}}}\,. 
\end{equation}
\end{Lemma}
{\bf Proof:}  (\ref{eq:AronsKrein}) is a special case of the so-called 
Aronszajin-Krein formula \cite{Simon,SimonTrace,Kiselev,Albeverio2004,Albeverio1997}. It can be easily seen as follows.
Set $A_z:=A-z$. Moreover, by slightly abusing notation define
the linear form $\bc^T:=\langle \bc,.\rangle$.
Using the
Sherman-Morrison formula \cite{bartlett1951}
\[
(A_z+\bc\bc^T)^{-1}= A_z^{-1} -\frac{A^{-1}_z \bc\bc^T A^{-1}_z}{1+ \langle \bc,A^{-1}_z \bc\rangle}\,,
\]
one easily obtains
\[
\langle \bc,(A_z +\bc\bc^T)^{-1} \bc\rangle =   \frac{\langle \bc,A_z^{-1} \bc\rangle}{1+  
\langle \bc,A_z^{-1} \bc\rangle}\,.
\]
Then the statement follows using
\begin{eqnarray*}
F_{\mu_{A,\bc}}(z) &=&   -\langle \bc, A_z^{-1} \bc\rangle \\
F_{A+cc^T,c}(z) &=& -\langle \bc, (A_z+\bc\bc^T)^{-1} \bc\rangle  \,.
\end{eqnarray*}
$\Box$

\vspace{0.2cm}
\noindent
By applying Lemma~\ref{lem:cauchywithbwithoutb} to the operator
$A_\nu$ in (\ref{eq:Tdefgeneral}) we obtain:
\begin{Corollary}[Cauchy transform of rank one perturbation]
For any finite measure $\nu$ on $\R$, 
the Cauchy transforms of $\nu$ and $R(\nu)$ are related by
\[
F_{R(\nu)}= \frac{F_\nu}{1- F_\nu}\,.
\]
\end{Corollary}
Moreover, we will need the following map:
\begin{Definition}[multiplication map]
If $\nu$ denotes a Borel measure on $\R$, we define
$M(\nu)$ by
\[
\int f(\lambda) d M(\nu) (\lambda) =\int f(\lambda) \lambda^{-2} d\nu(\lambda),
\] 
for every measurable function $f$ on $\R$.
\end{Definition} 
The transformation $M$ describes 
how the spectral measure induced by $A$ and $\psi$ changes when $\psi$ is replaced with $A^{-1}\psi$, that is
\[
M(\mu_{A,\psi})= \mu_{A,A^{-1}\psi}.
\] 
This is also easily verified by diagonalizing $A$
to a multiplication operator on $L^2(\R,\mu_{A,\psi})$
as above. 

Using the transformations $M$ and $R$, we obtain the following concise form for the spectral measure induced by the confounding vector:
\[
\mu_{\sx,c\sx^{-1}\bb}=c^2 \mu_{\sx,\sx^{-1}\bb}=
c^2 \mu_{(\se+\bb\bb^T),(\se+\bb\bb^T)^{-1} \bb}
=c^2 M[R[\mu_{\se,\bb}]]. 
\]

We will also need the following result:
\begin{Lemma}[weak continuity of $R$ and $M$ ]\label{lem:rmcontinuous}
Let $\nu_d$ be a sequence of measures with common support $[l,r]$ with $r>l>0$. 
If $\nu_d \to \nu$ weakly
then
$
R(\nu_d) \to R (\nu)
$
and
$M(\nu_d) \to M(\nu)$ weakly.
\end{Lemma}
{\bf Proof:} 
Since $\nu_d$ converges weakly to $\nu$,
$F_{\nu_d}$ converges pointwise to $F_\nu$.
Thus, $F_{R(\nu_d)}$ converges pointwise
to $F_{R(\nu)}$ for all $z\in \C^+$. Due to Theorem~10 in
\cite{MingoSpeicher}, $F_{R(\nu)}$ is the Cauchy transform
of the limit of $R(\nu_d)$,
see also \cite{Bercovici}, Section~5.

Weak continuity of $M$ is immediate since $M$ is the multiplication with a function that is bounded by 
$1/r^2$ and $1/l^2$.
$\Box$

\vspace{0.2cm}
\noindent

Since we observe $\Sigma_{\bX\bX}$ and not 
$\Sigma_{\bE\bE}$ is is important for our
purpose that the tracial measure of both matrices
asymptotically coincide. The infinite version of Lemma~\ref{lem:tracialclose}
reads:
\begin{Lemma}[tracial measures coincide]\label{lem:tracialequal}
If $\mu_{\sx^d,\tau}\to \mu^\infty$ 
weakly then $\mu_{\se^d,\tau}\to \mu^\infty$ weakly, too.
\end{Lemma}
{\bf Proof:} We have for every interval $[r,l]$:
\begin{eqnarray*}
& &\lim_{d\to\infty}  |\mu_{\sx,\tau}[r,l]-\mu^\infty[r,l]| \leq\\
 &&
\lim_{d\to\infty} |\mu_{\sx,\tau}[r,l]-\mu_{\se,\tau}| +
\lim_{d\to \infty} |\mu_{\se,\tau} [r,l]- \mu^\infty [r,l]| \,.
\end{eqnarray*}
The first term is zero due to Lemma~\ref{lem:tracialclose} and the second one by assumption. 
Since the intervals generate the entire Lebesgue Borel sigma algebra the statement follows.
$\Box$

\vspace{0.2cm}
\noindent
To derive the asymptotic for the confounding strength $\beta_d$ we observe
\[
\beta= \frac{ c^2 M[R[r_\bb^2 \mu^\infty]](\R)}{
r^2_\ba +  c^2 M[R[r_\bb^2 \mu^\infty]](\R)}
\]

We are now prepared to prove Theorem~\ref{thm:relgenericfamily}.
Due to Theorem~\ref{thm:justi} we have
\begin{eqnarray*} 
\lim_{d\to\infty } \mu_{\sx^d,\ba_d+c (\sx^d)^{-1} \bb_d}
&=& \lim_{d\to \infty} \mu_{\sx^d, \ba_d }
+  \lim_{d\to \infty} \mu_{\sx^d,c (\sx^d)^{-1} \bb_d}
= r^2_\ba \mu^\infty + c^2 \mu_{\sx^d, (\sx^d)^{-1} \bb_d}
\\
&=&  r^2_\ba \mu^\infty +  c^2 \lim_{d\to \infty}  M[R[\mu_{\se^d, \bb_d}]].
\end{eqnarray*}
Due to Lemma~\ref{lem:rmcontinuous} and   
Theorem~\ref{thm:justi}
we have
\[
\lim_{d\to \infty}  M[R[\mu_{\se^d, \bb_d}]]=
M[R[r_\bb^2 \mu^\infty]].  
\]
Hence we obtain
\begin{equation}\label{eq:muahatlim}
\lim_{d \to \infty} \frac{\mu_{\sx^d,\hat{\ba}_d}}{\|\hat{\ba}_d\|^2}=
\frac{r^2_\ba \mu^\infty + c^2 M[R[r_\bb^2 \mu^\infty]]}{\lim_{d\to \infty} \|\hat{\ba}_d\|^2}.
\end{equation}
To evaluate the denominator on the right hand side, we employ
\eqref{eq:normalization} and obtain:
\begin{eqnarray*}
\lim_{d\to \infty} \|\hat{\ba}_d\|^2 &=&
\lim_{d \to \infty} \mu_{\sx^d,  \ba_d + c (\sx^d)^{-1}\bb_d } (\R) 
=
\lim_{d\to \infty} \mu_{\sx^d, \ba_d} (\R) + 
\lim_{d\to \infty} \mu_{\sx,c (\sx^d)^{-1} \bb_d} (\R)\\
&=& r_\ba^2 \mu^\infty(\R) + \lim_{d\to\infty} c^2 M[R[\mu_{\se^d,\bb_d}]](\R) = r^2_\ba +   
c^2 M[R[r^2_\bb \mu^\infty]](\R). 
\end{eqnarray*}
Inserting this into \eqref{eq:muahatlim} yields:
\begin{equation}\label{eq:muahatlimfinal}
\lim_{d\to \infty} \frac{\mu_{\sx^d,\hat{\ba}_d}}{\|\hat{\ba}_d\|^2} 
= \frac{r^2_\ba }{r^2_\ba +   
c^2 M[R[r^2_\bb \mu^\infty]](\R)}  \mu^\infty +
\frac{c^2}{r^2_\ba +   
c^2 M[R[r^2_\bb \mu^\infty]](\R) }    M[R[r^2_\bb \mu^\infty]].
\end{equation}
 On the other hand, recalling the construction of $\nu_{\beta,\eta}$ in Section~\ref{sec:asym},
for fixed $d$ (which we drop first) we have 
\begin{eqnarray*}
\nu_{\beta,\eta}  &=&
 (1-\beta) \mu_{\sx,\tau} + \beta \frac{\mu_{T,T^{-1} \bg}}{\|T^{-1}\bg\|^2} =
 (1-\beta) \mu_{\sx,\tau} + \beta \frac{M[\mu_{T,\bg}]}{M[\mu_{T,\bg}](\R)}\\
 &=&  (1-\beta) \mu_{\sx,\tau} + \beta \frac{M[R[\mu_{M_X,\bg}]]}{M[R[\mu_{M_X,\bg}]](\R)}
 =  (1-\beta) \mu_{\sx,\tau} + \beta \frac{M[R[\mu_{\sx,\tau}]]}{M[R[\mu_{\sx,\tau}]](\R)}
\end{eqnarray*} 
Hence we obtain: 
\begin{eqnarray*}
\lim_{d\to \infty} \nu_{\beta,\eta}^d  &=&
\lim_{d\to \infty} (1-\beta) \mu_{\sx^d,\tau} +
\lim_{d\to \infty} \beta \frac{M[R[\eta \mu_{\sx^d,\tau}]]}{M[R[\eta \mu_{\sx^d,\tau}]](\R)}\\
&=& \frac{r^2_\ba}{c^2 M[R[r^2_\bb \mu_{\sx^d,\tau}]](\R) +r^2_\ba} +\frac{c^2}{
c^2 M[R[r^2_\bb \mu_{\sx^d,\tau}]](\R)+r^2_\ba} 
M[R[r^2_\bb \mu_{\sx^d,\tau}]],
 \end{eqnarray*}
which coincides with \eqref{eq:muahatlimfinal}.

\section{Experiments with simulated data}

\subsection{Estimation of strength of confounding}

\label{subsec:expestimatebeta}

We first ran experiments where the data has been generated according to our model assumptions:
First, both the influence of $\bX$ on $Y$ and the influence 
of $Z$ on $\bX$ and $Y$ is linear. Second, the vectors
$\ba$ and $\bb$ are randomly drawn from a uniform
distribution over the unit sphere. More specificly,
the data generating process reads as follows:

\begin{itemize}
\item {\bf Generate $\bE$:} first generate $n$ samples
of a $d$-dimensional vector valued Gaussian random variable $\tilde{\bE}$ with mean zero and covariance matrix ${\bf 1}$.
Then generate a random matrix $G$ whose entries are independent standard Gaussians and set $\bE := G \tilde{\bE}$.

\item {\bf Generate scalar random variables $Z$ and $F$} by drawing $n$ samples of each independently from a standard Gaussian distribution.

\item {\bf Draw scalar model parameters $c,r_\ba,r_\bb$} by independent draws from the
uniform distribution on the unit interval.

\item {\bf Draw vectors $\ba,\bb$} independently 
from a sphere of radius $r_\ba$ and $r_\bb$, respectively. 

\item {\bf Compute $\bX$ and $Y$} via the structural equations  \eqref{eq:xcon}. 

\end{itemize}

Note that for the above generating process
the computation of the true confounding strength $\beta$ involves only model parameters that are exactly known, even $\sx$ need not be estimated from the data matrix because it is simply  given by $GG^T +\bb \bb^T$.

\begin{figure}
\centerline{
\begin{tabular}{ccc}
\includegraphics[width=0.33\textwidth]{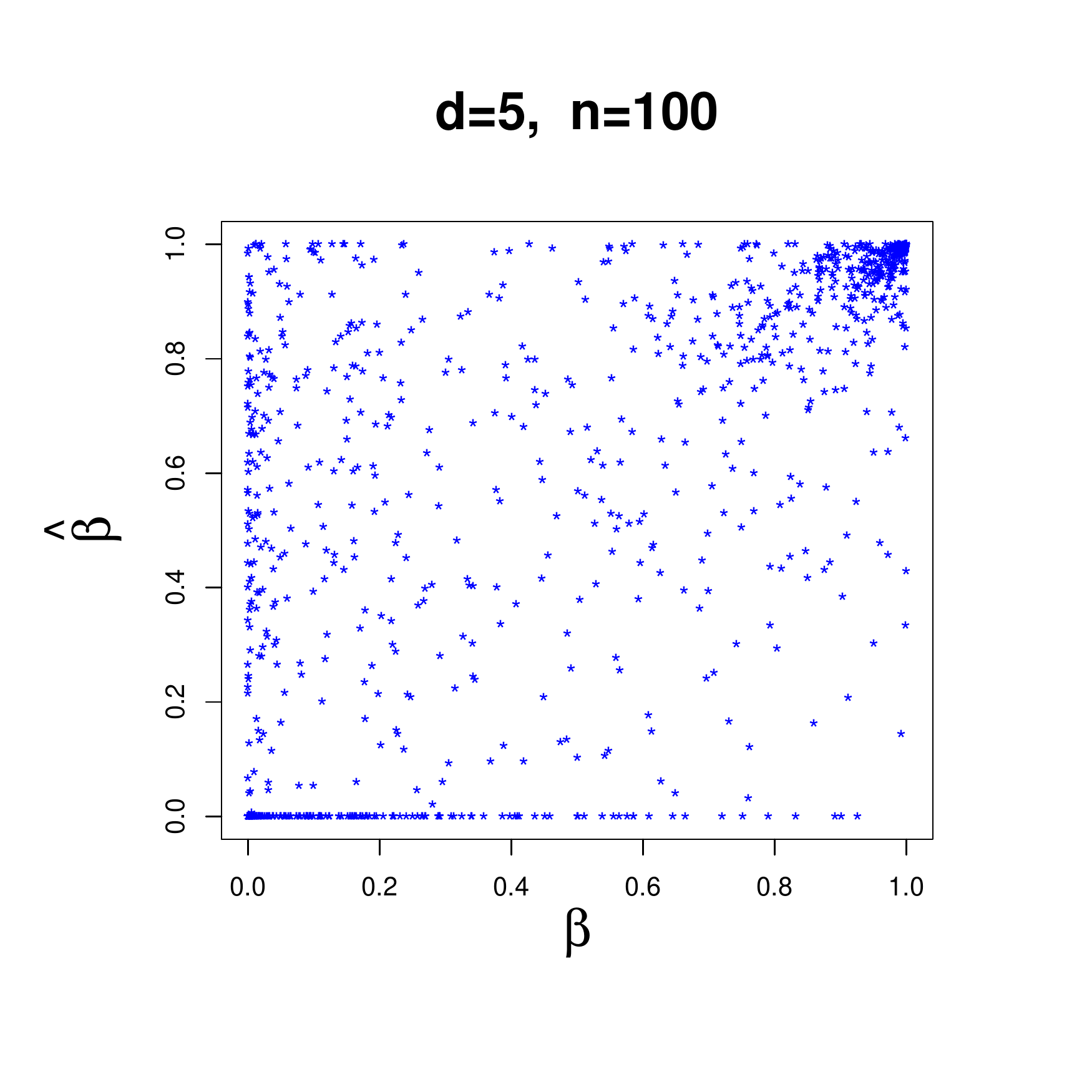}   &
\includegraphics[width=0.33\textwidth]{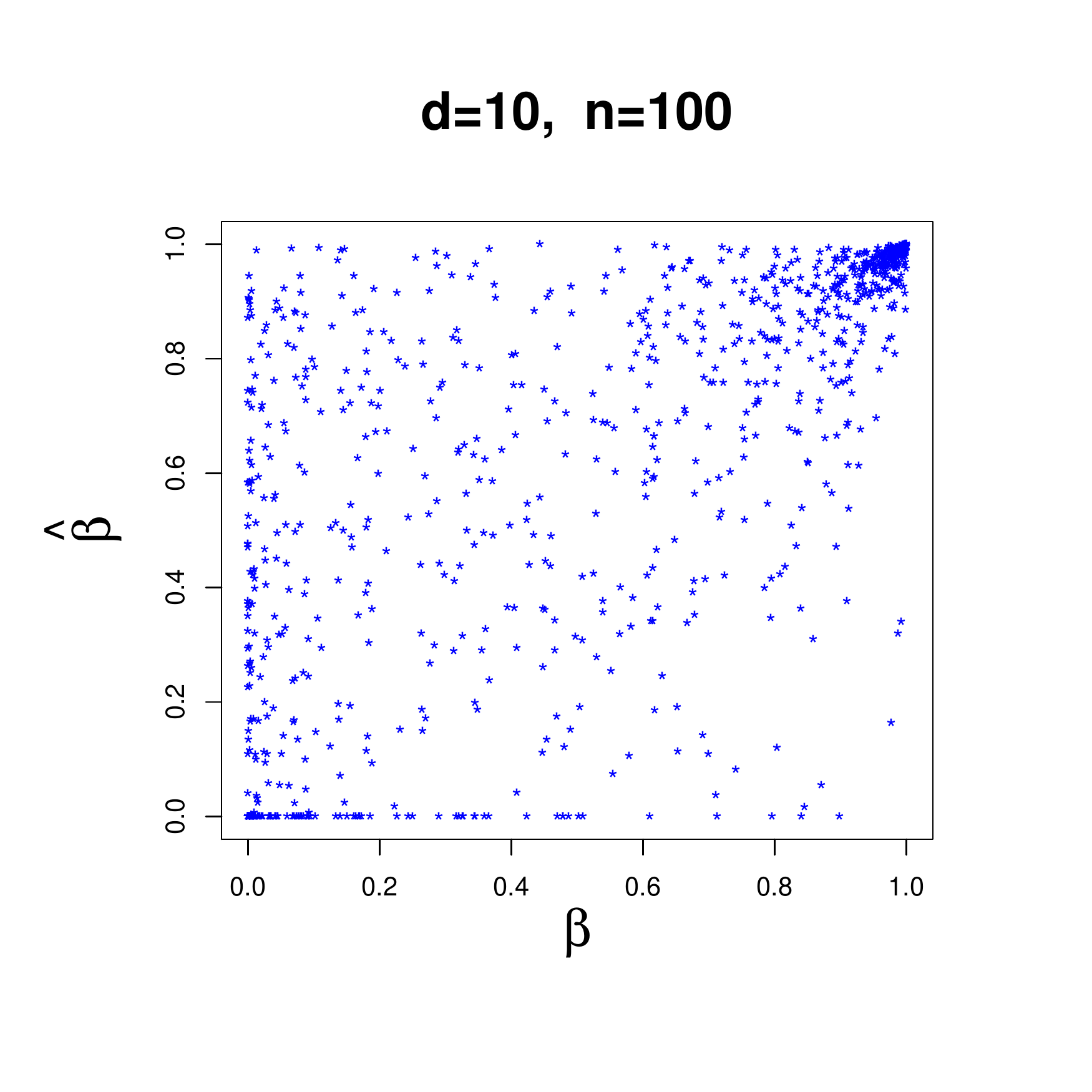} &
\includegraphics[width=0.33\textwidth]{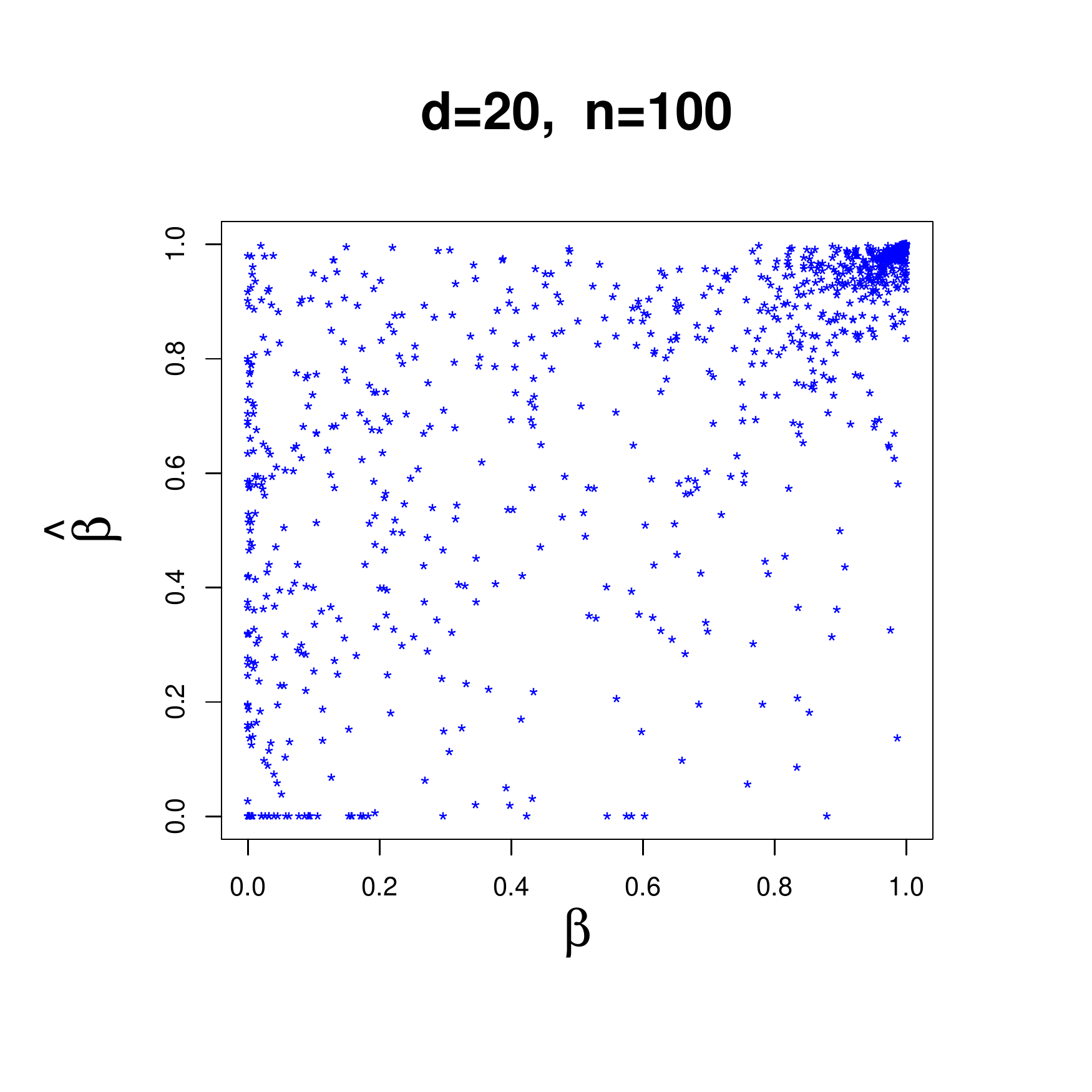} 
\\
\includegraphics[width=0.33\textwidth]{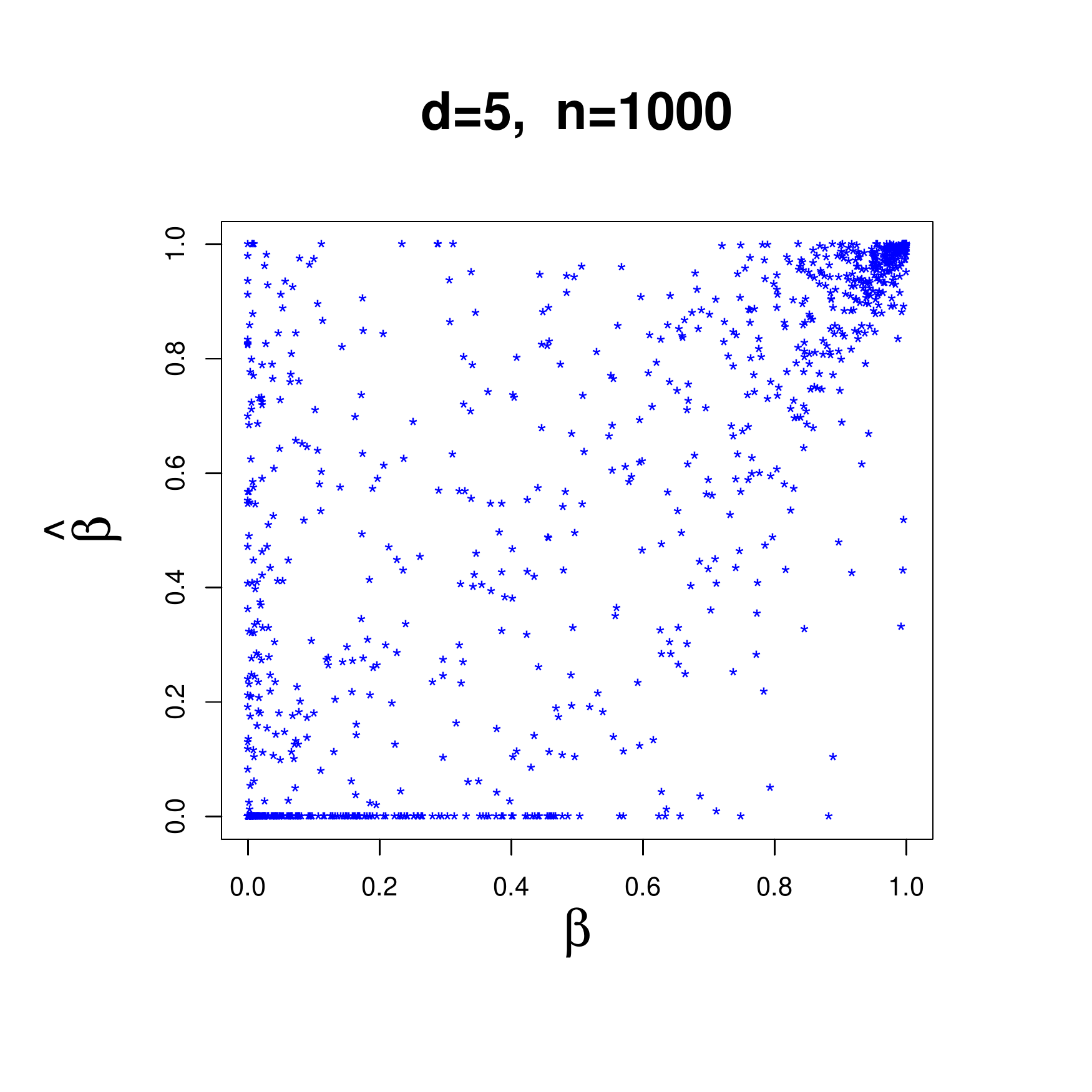}  &
\includegraphics[width=0.33\textwidth]{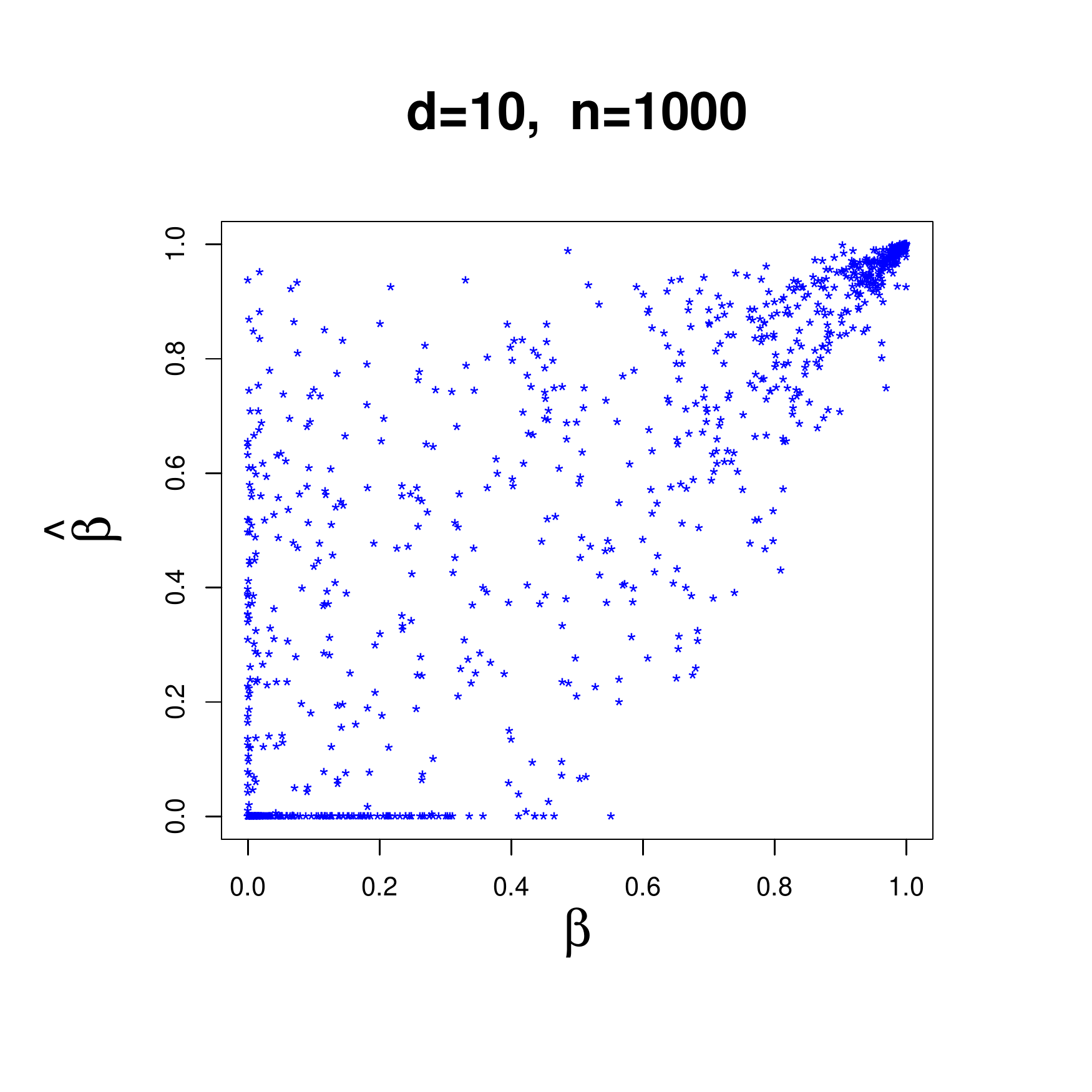}&
\includegraphics[width=0.33\textwidth]{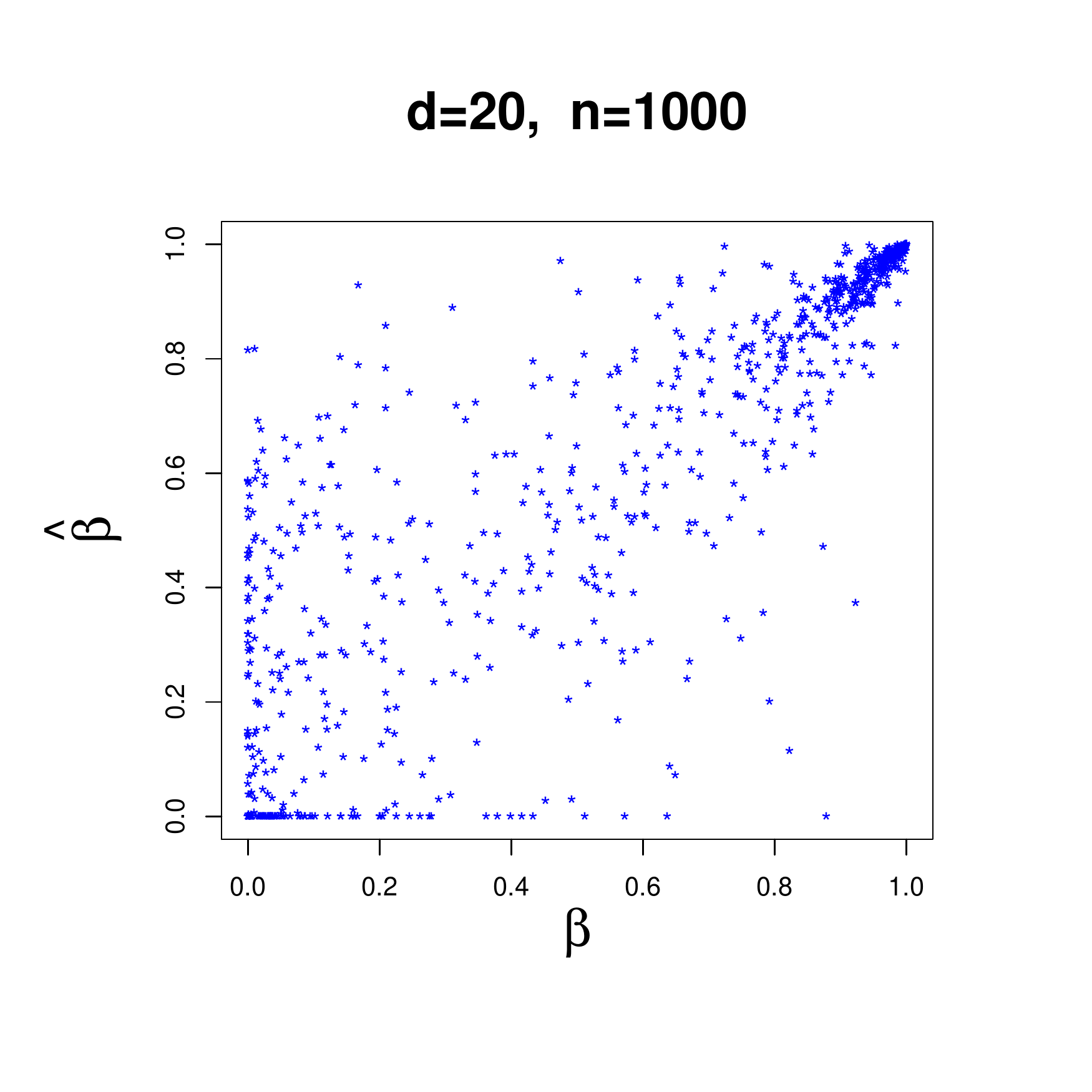} 
\\                            
 \includegraphics[width=0.33\textwidth]{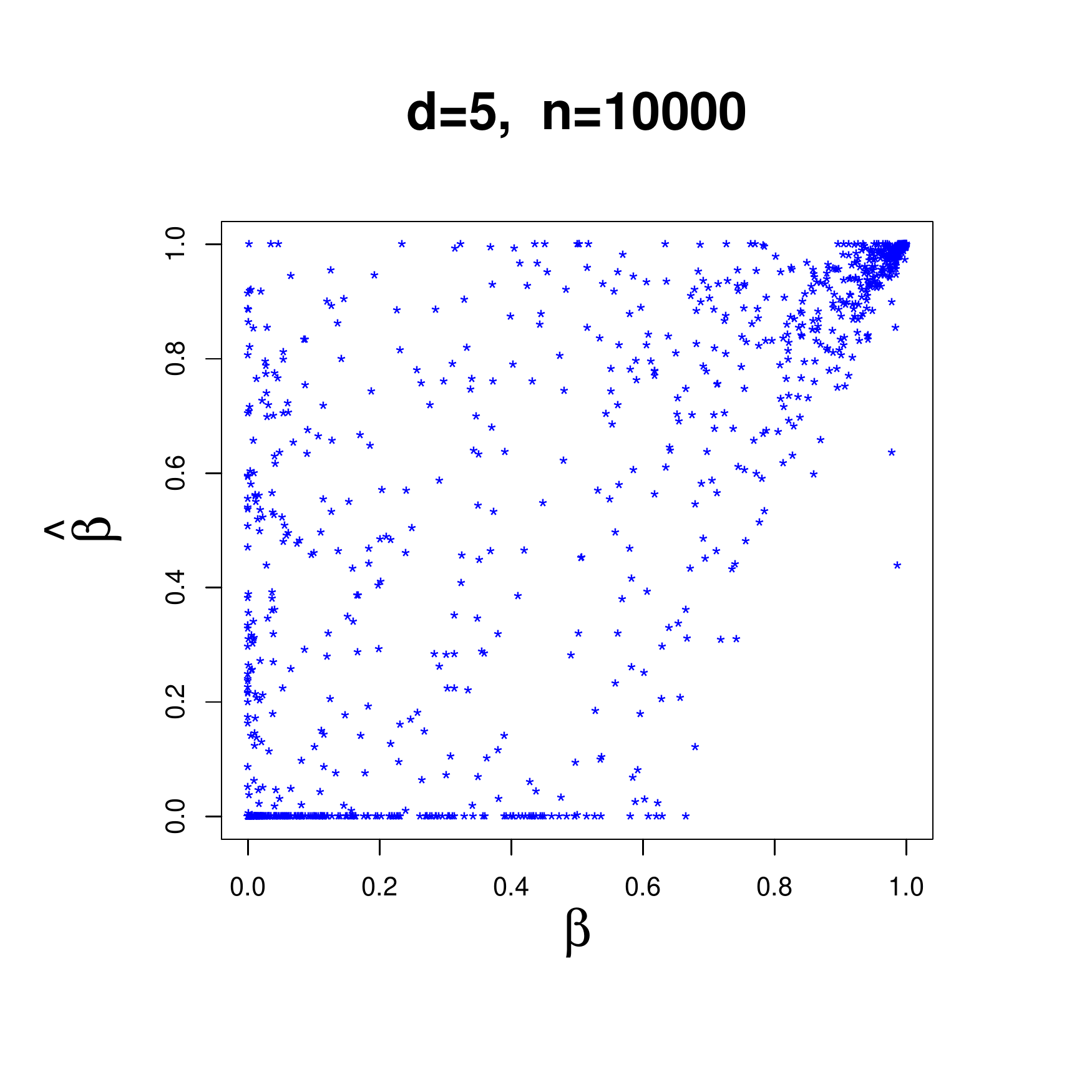} &
\includegraphics[width=0.33\textwidth]{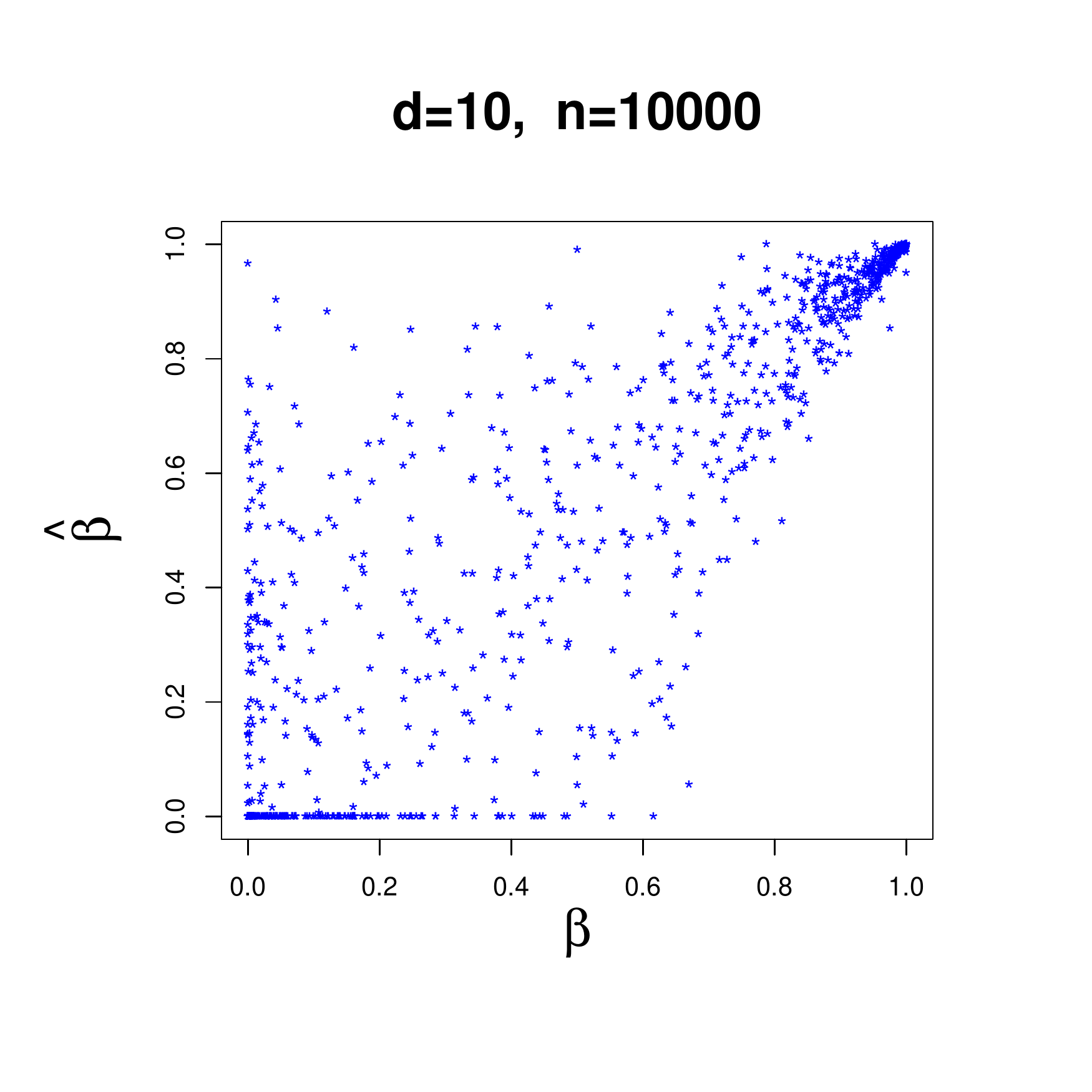} &
\includegraphics[width=0.33\textwidth]{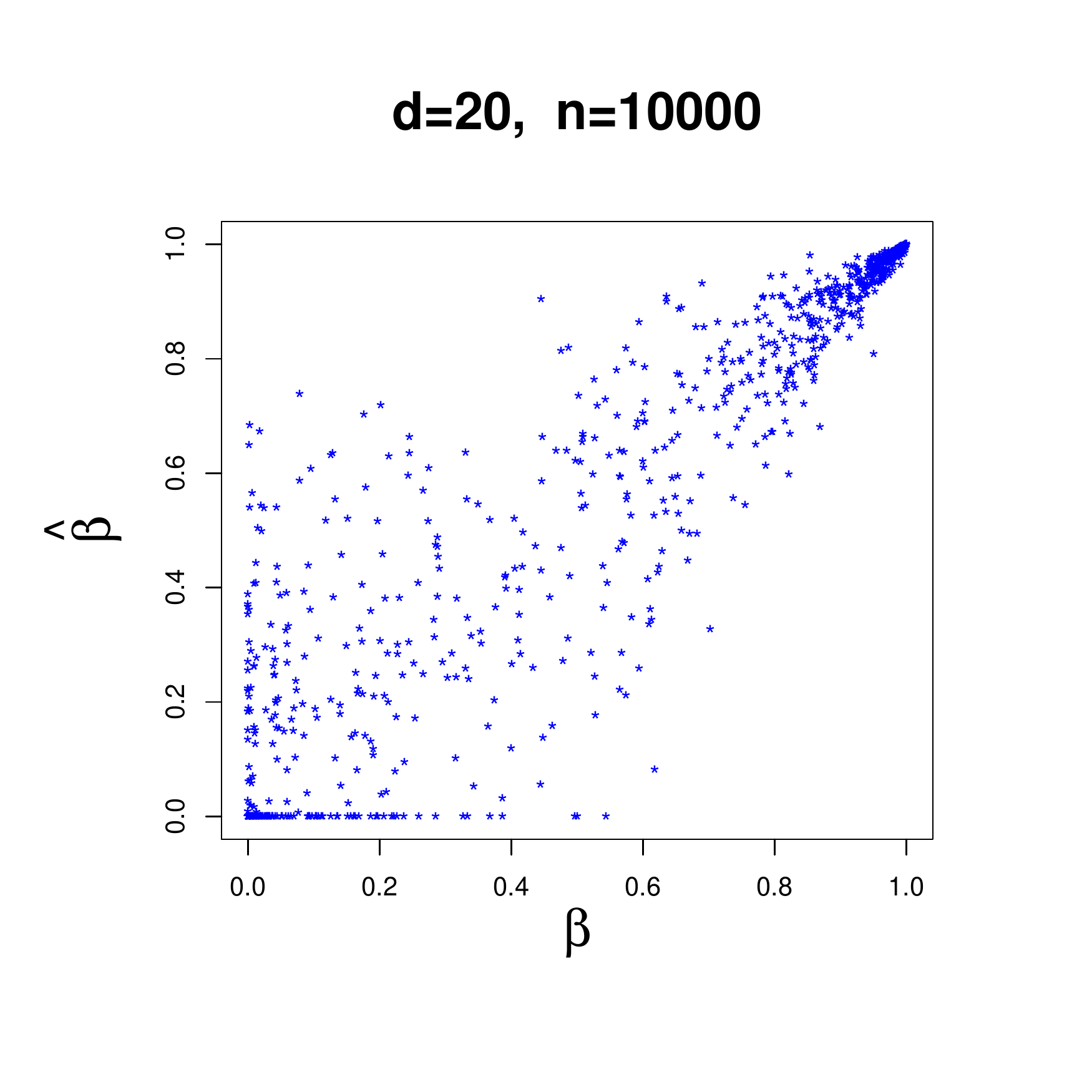} 
\\ 
\includegraphics[width=0.33\textwidth]{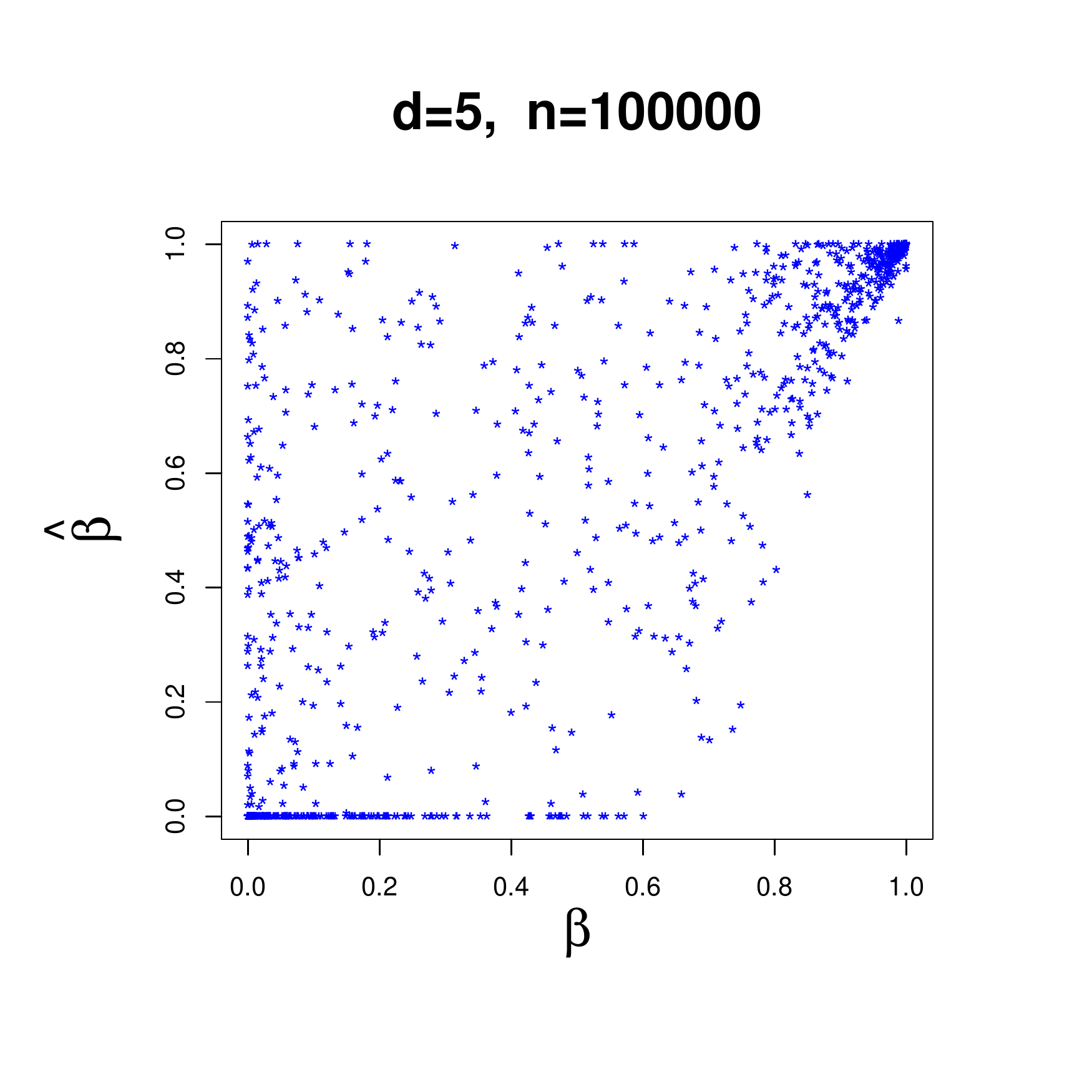}   &
\includegraphics[width=0.33\textwidth]{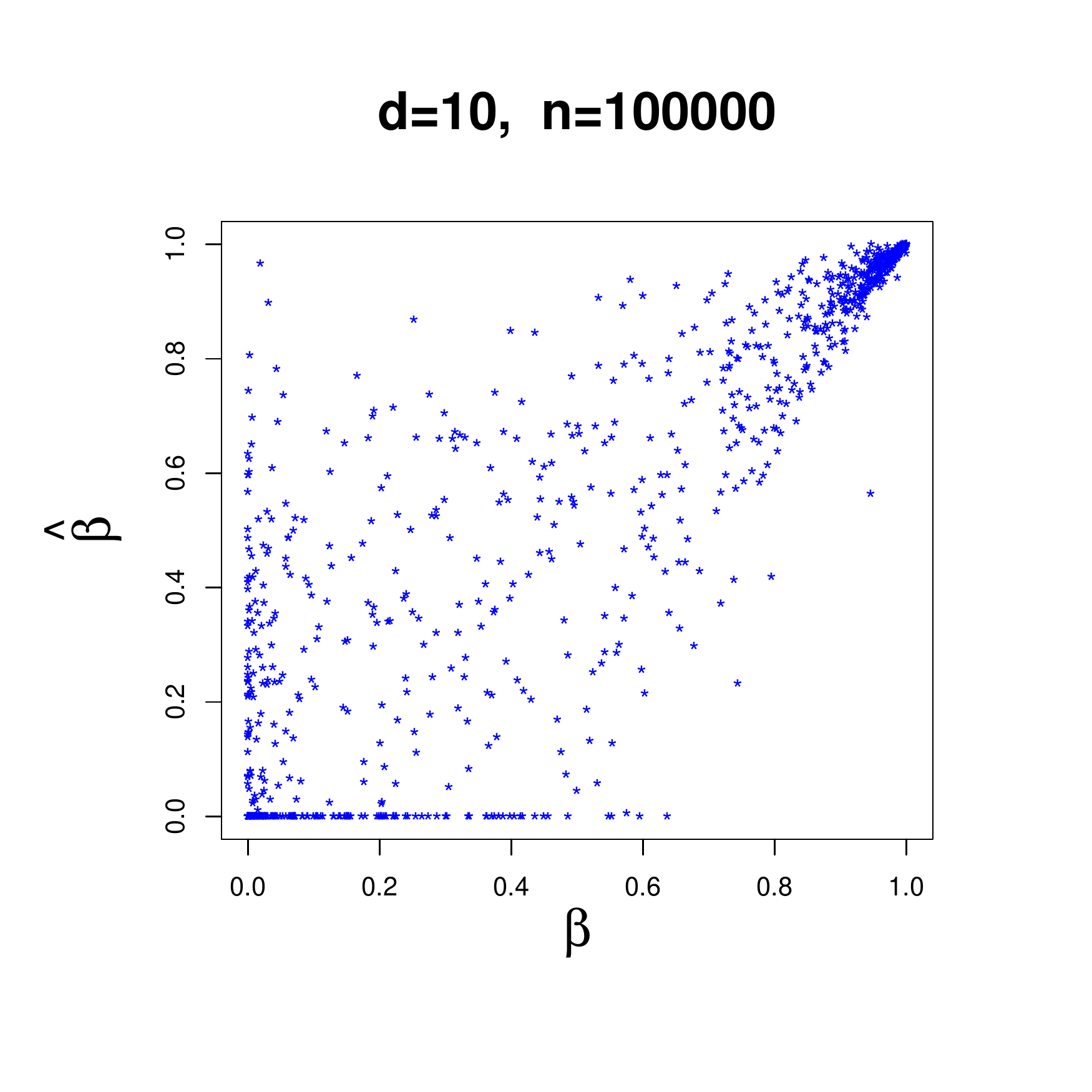}&
\includegraphics[width=0.33\textwidth]{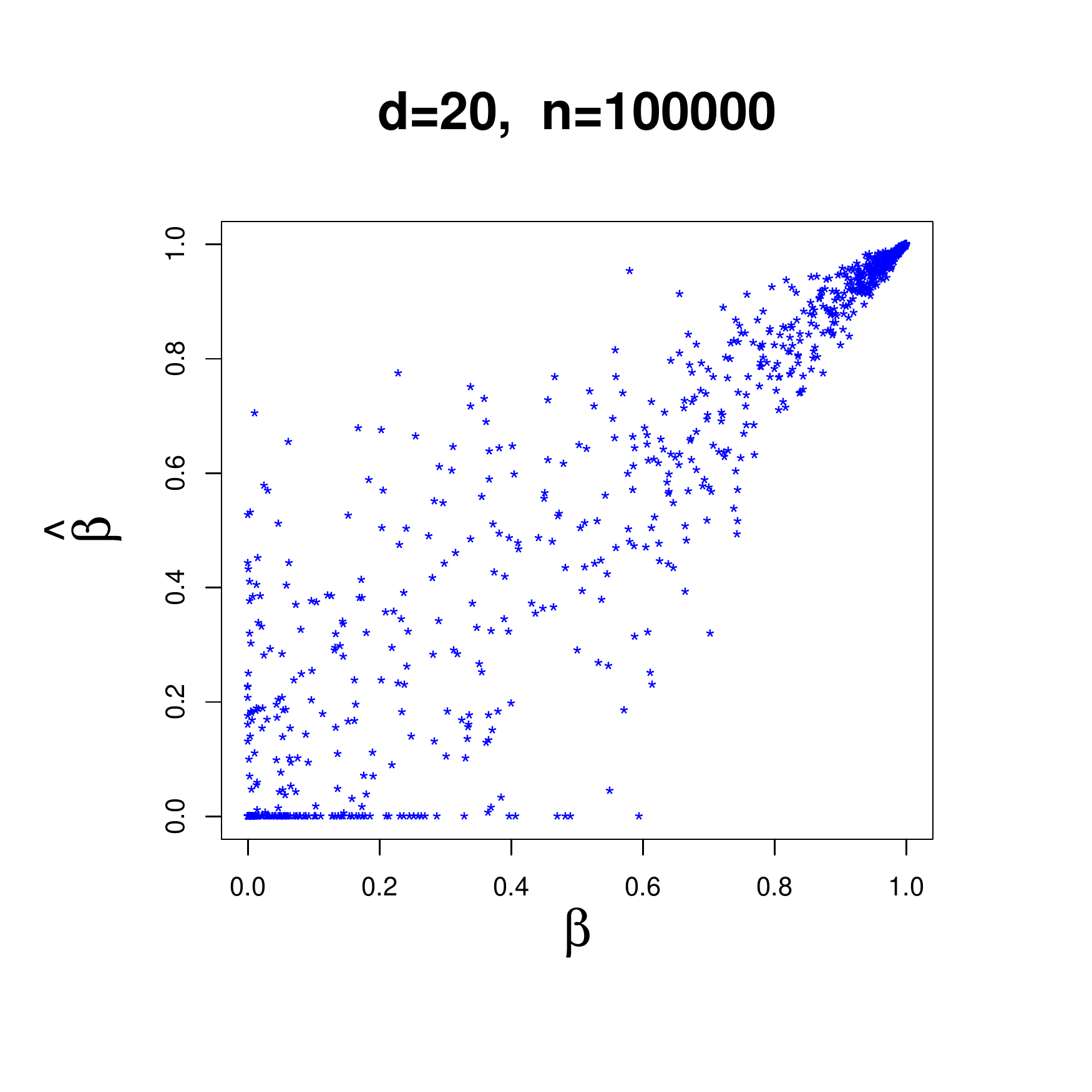}    
\end{tabular}
}
\caption{\label{fig:allsimulations} Simulation results: true value $\beta$ versus estimated 
value $\hat{\beta}$ for different dimensions $d$ and sample sizes $n$. The performance inreases with higher dimensions provided that the sample size is large enough.} 
\end{figure}

Figure~\ref{fig:allsimulations} shows the results for dimensions $d=5;10;20$ and
sample sizes $n=100; 1000; 10,000; 100,000$.
They indicate that the sample size is even more critical for the performance than
the dimension. It seems that the required sample sizes grow so quickly with the dimension
that data sets with sample sizes under  $1000$ should 
only be considered if the dimension is not larger than
about $10$. 

Although the results for dimension $5$ look quite bad, it should be noted that
$\beta$ and $\hat{\beta}$ are already significantly correlated,
the correlations coefficients 
varied in the range between $0.65$ and $0.8$ for the different sample sizes
with $p$-values below $10^{-10}$.

We found the true and estimated value of $\eta$ to be quite uncorrelated,
it seems that $\eta$ is hard to estimate using our method. 
Since our focus is on the confounding strength, we will not explore this any further.

\section{Experiments with real data under controlled conditions}

It is hard to find real data where the strength of confounding is known.
This is because there are usually unknown confounders in addition to the ones
that are obvious for observers with some domain knowledge.
For this reason, we have designed an experiment where the variables are observables of technical devices among which the causal structure is known by construction of the experimental setup.

\subsection{Setup for a confounded causal influence}

\begin{figure}
\centerline{
\includegraphics[width=\textwidth]{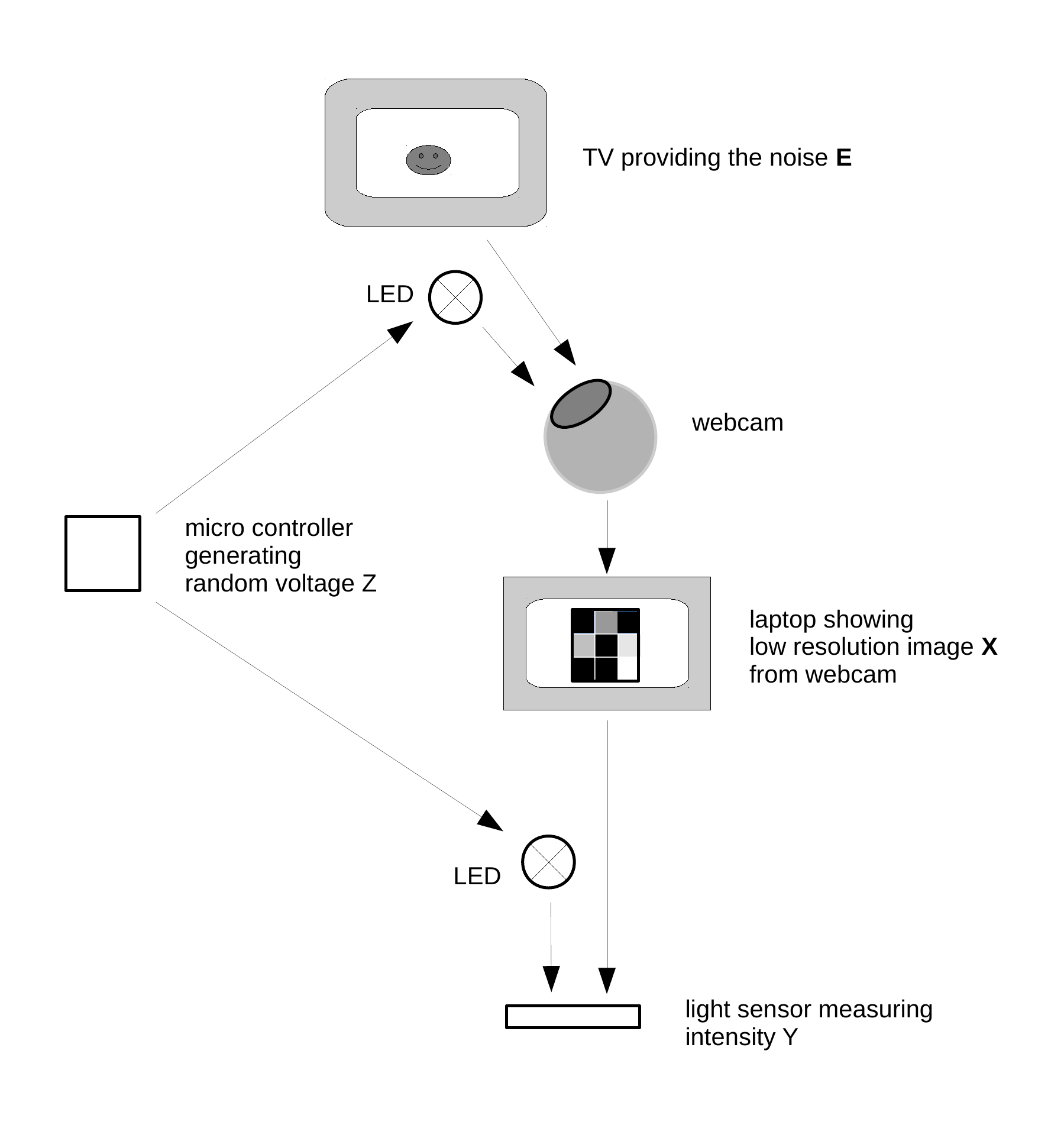}
}
\caption{\label{fig:generic_setup} Setup for the generic causal relation 
where $\bX$ influences $Y$ and $Z$ influences both $\bX$ and $Y$, see text. Note that we used the symbol for light bulbs to represent 
the LEDs in order to simplify the drawing.}  
\end{figure}

\begin{figure}
\centerline{
\includegraphics[width=0.6\textwidth]{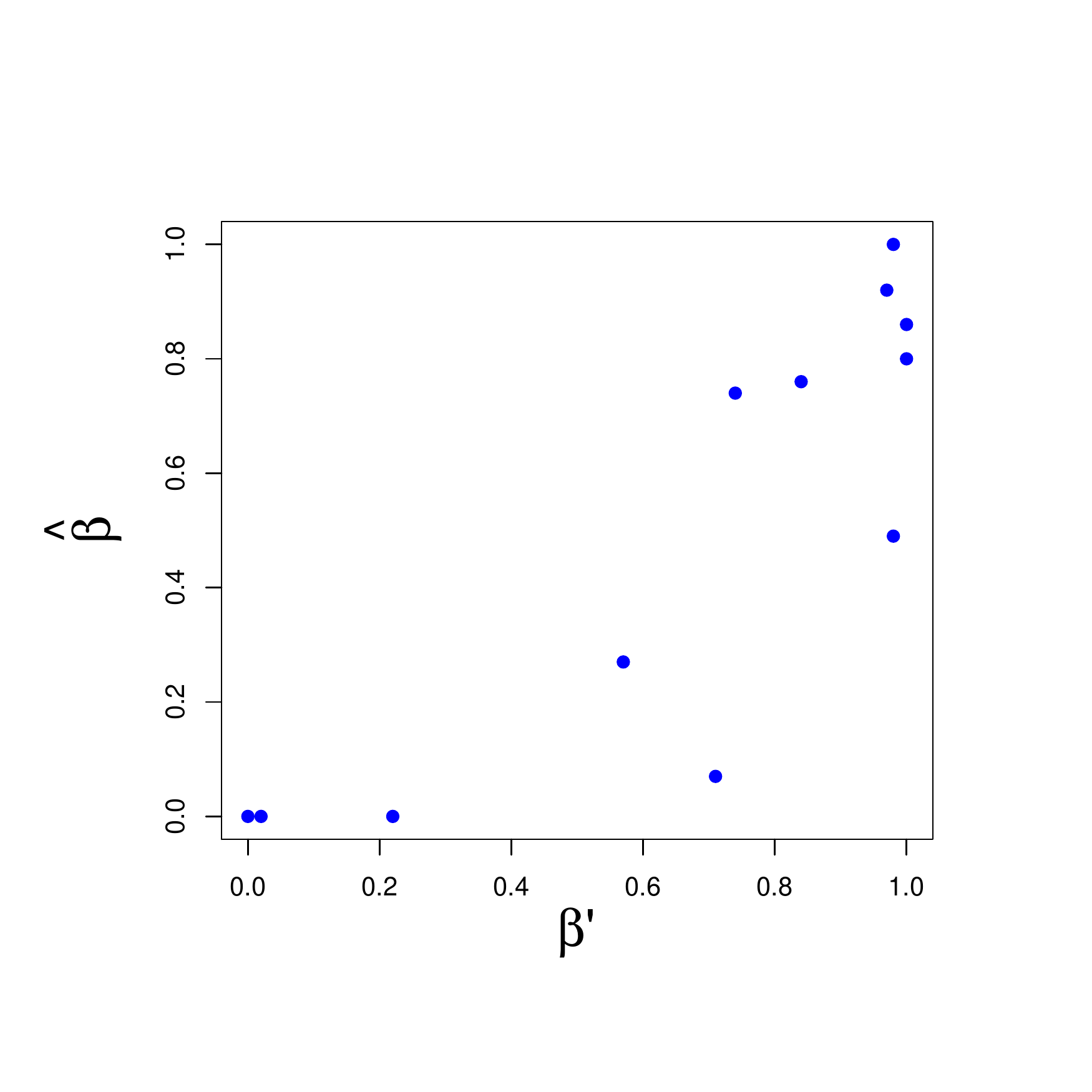}
}
\caption{\label{fig:confounded_brightness} Confounding strength estimated using observations of $Z$ vs. confounding strength estimated by our algorithm for sample size $1000$.} 
\end{figure}

To obtain a causal relation where $\bX$ influences $Y$ and there is, a the same time, a confounder $Z$ influencing both $\bX$ and $Y$, we have chosen the setup shown in Figure~\ref{fig:generic_setup}.\footnote{The dataset will be made available online after acceptance.}
The cause $\bX$ is a $9$-dimensional pixel vector generated by extremely reducing the resolution of an image taken by a webcam to $3\times 3$ pixels. The effect 
$Y$ is the intensity measured at a light sensor in front of a laptop screen that displays the $3\times 3$ image, amplified to a size of about $10\times 10$ centimeter. The sensor 
is located at a distance of about $10$ centimeter from the screen. 
To confound the causal relation by a common cause $Z$, we have generated an independent random  voltage that controls the brightness of two LEDs: one influencing $\bX$ 
because it is placed in front of the webcam and one that is placed in front of the light sensor. 
To ensure that $\bX$ is not entirely determined by the LED,
we have placed the webcam in front of a TV. This way, 
the image taken by the webcam is influenced by both the LED and  the TV signal -- the latter plays the role of
$\bE$ in our structural equation \eqref{eq:xcon}. 
To avoid that fluctuations of daylight is an additional confounder we have covered the pair sensor and laptop screen by a towel. Since we have measured $Z$ (the random value of the voltage which determines the brightness of the LEDs),
we are able to compute the strength $\beta$ of confounding up to an extent where the estimations of $\sx, \bb, \ba$ 
from empirical data coincide with their true counterparts.
We will denote this value by $\beta'$ to emphasize that
it may still deviate from the true value $\beta$ when the sample size is not sufficient. 
Figure~\ref{fig:confounded_brightness} shows $\beta'$ and $\hat{\beta}$ for experiments with sample size $1000$.
For this setup, the algorithm tends to underestimate confounding, but shows qualitatively the right tendency
since $\beta'$ and $\hat{\beta}$ clearly correlate.

After inspecting some spectral measures for the above scenario
we believe that the algorithm underestimates confounding for the following reason: 
The vector $\ba$ describing the influence of the images
on the sensor is not in generic orientation relative to $\sx$. This is because the pixels are usually {\it positively} correlated
and each pixel has {\it positive} influence on the total intensity measured by the sensor. This way, $\ba$ has stronger weights for high eigenvalues. Since confounding often increases the weights of low eigenvalues (note, however, that this depends 
also on our second confounding parameter $\eta$), the ''non-genericness'' of the vector $\ba$ tends to compensate the effect 
of confounding on the spectral measure induced by $\hat{\ba}$.
It is likely that such an underestimation of confounding occurs for many other scenarios as well. This is because it is not uncommon that all variables in a vector are positively correlated\footnote{Note also the concept of multivariate total positivity of order two (MTP2) \cite{Karlin80}, which implies positive correlations between all variables and occurs in many applications \cite{Fallat16} such as Markov random fields that consist only of attractive interaction terms.} and that they all have a positive influence on some target variable.

The above setting contained the purely confounded and the 
purely causal scenario as limiting cases:
the confounded one by putting the LEDs close to the sensor and
the webcam, respectively, and setting the voltage to the maximal values,
and the purely causal one  by setting the voltage to zero.

To get further support for the hypothesis that the algorithms
tends to behave qualitatively in the right way even when the
estimated strength of confounding deviates from its true value,
we also tested modifications of the above scenario that
are purely confounded or purely causal {\it by construction} and not only by variations of parameters. This is described in
Sections~\ref{subsec:purelycausalimage}    and \ref{subsec:purelycausalimage}.

\subsection{Purely causal scenarios \label{subsec:purelycausalimage}}

To build a scenario without confounding, $\bX$ is the pixel vector of
grey values of an image section (consisting  of $3\times 3$ pixel)
randomly drawn from a fixed image. 
The image sections are displayed by the screen of a laptop, amplified to a size
of about $10$\,cm $\times 10$ \,cm. A light sensor, placed in front of the screen
with a distance of about $10$\, cm, measures the light intensity, which is our variable $Y$. Clearly, $\bX$ influences $Y$ in an unconfounded way because 
the selection of the images is perfectly randomized. Fluctuations of the brightness caused by the environment certainly influence $Y$, but count as independent noise since they do not influence $\bX$. 

We tried this experiment with sample size $1000$, where the estimated confounding strength was $\hat{\beta}=0$.  We should also mention that we obtained $\beta'=6.8 \cdot 10^{-7}$
in agreement with our statement that the experimental setup is unconfounded
because the image sections are drawn randomly. The extremely low value of $\beta'$ also shows that $\beta'$ is indeed very close to the true value $\beta$, which justifies to identify them.

\subsection{Purely confounded scenario\label{subsec:purelyconfoundedimage} }

The setup in Figure~\ref{fig:generic_setup} can be easily modified to a causal structure
where the relation between the pixel vector $\bX$ and the light intensity $Y$ is purely confounded by a one-dimensional variable $Z$: we just need 
to put the light sensor to a place where it neither sees the TV nor the screen of the laptop. If we, again, ensure that  the light sensor is not influenced by the same fluctuations of daylight as the webcam (e.g. by  covering the sensor by a towel), the 
statistical dependence between $\bX$ and $Y$ is due to $Z$, that is, the fluctuations of the random
light signal from the LEDs alone. 

We have performed this experiment with sample size $1000$ and obtained
$\hat{\beta}=0.68$ and $\beta'=0.998$, which again is consistent with 
our previous observation that confounding is underestimated.

\section{Experiments with real data with partially known causal structure}

The experiments in this section refer to real data where the
causal structure is not known with certainty. For each data set, however, we will briefly discuss the plausibility of the results
in light of our limited domain knowledge. The main purpose of the section is to show that the estimated values
of confounding strength  indeed spread over the whole interval $[0,1]$. A priori, we could not be sure whether  empirical data follow probability distributions that are so different from our
model assumptions that only small or only large values of confounding were estimated.

\subsection{Taste of wine}

This dataset  \cite{Lichman2013} describes the dependence between $Y$,
the scores on the taste between 0 and 10  (given by human subjects) of red wine, and 11 different ingredients:
$X_1$: fixed acidity, 
$X_2$: volatile acidity, 
$X_3$: citric acid, 
$X_4$: residual sugar, 
$X_5$: chlorides, 
$X_6$: free sulfur dioxide, 
$X_7$: total sulfur dioxide, 
$X_8$: density, 
$X_9$: pH, 
$X_{10}$: sulphates, 
$X_{11}$: alcohol.  
It turned out that the largest eigenavlue of the covariance matrix
is by orders of magnitude larger than the others. We therefore 
normalized the $X_j$ to unit variance and obtained the
covariance matrix  
\[
\sx=\left(\begin{array}{ccccccccccc}
 1 & -0.26 & 0.67 & 0.11 &  0.09 & -0.15 & -0.11 &  0.67 & -0.68  &0.18 &-0.06\\ 
-0.26 & 1.00 &-0.55 & 0.00 & 0.06 &-0.01 &  0.08 &  0.02 &  0.23 &-0.26 &-0.20\\  
  0.67 & -0.55 & 1 &  0.14 &  0.20 & -0.06 &  0.04 &  0.36 & -0.54 & 0.31 & 0.11\\  
 0.11 & 0.00 & 0.14 & 1 & 0.06 &  0.19 &  0.20 &  0.36 & -0.09 & 0.01 & 0.04\\
 0.09 & 0.06 & 0.20 & 0.06 &  1 &  0.01 & 0.05 &  0.20 & -0.27 & 0.37 &-0.22\\
  -0.15 & -0.01 & -0.06 & 0.19 & 0.01 & 1 &  0.67 & -0.02 &  0.07 & 0.05 &-0.07\\
  -0.11 &  0.08  & 0.04 &  0.20 &  0.05 &  0.67 &  1 &  0.07 & -0.07 & 0.04 &-0.21\\
  0.67 &  0.02 & 0.36 &  0.36 & 0.20 &-0.02 &  0.07 &  1 & -0.34 & 0.15 &-0.50\\
  -0.68 &  0.23 & -0.54 &-0.09 &-0.27 &  0.07 & -0.07 & -0.34  & 1  & -0.20 & 0.21\\
 
 0.18 & -0.26 & 0.31 & 0.01 & 0.37 &  0.05 & 0.04 &  0.15 & -0.20 & 1 &  0.09\\
-0.06 & -0.20 & 0.11 & 0.04 & -0.22 &-0.07 &-0.21 &-0.50 & 0.21 & 0.09 & 1\end{array}\right)
\]
We observe several correlation coefficients around $0.5$ and $0.6$, hence $\sx$ significantly differs
from the identity, which is important  because $\sx={\bf 1}$ would render teh method pointless.   
The vector of regression coefficients reads:
\[
\hat{\ba}= \tiny{(0.044, -0.194, -0.036, 0.023, -0.088,
 0.046, -0.107,-0.034, -0.064, 0.155, 0.294)}\,,
\] 
showing that alcohol has by far the strongest 
association with taste. According to common experience,
alcohol indeed has a significant influence on the taste. Also the other associations are likely to be mostly causal and not due to a confounder. 
We estimated confounding for this data set and
obtained $\hat{\beta}=0$.
\begin{figure}
\centerline{
\includegraphics[width=0.5\textwidth]{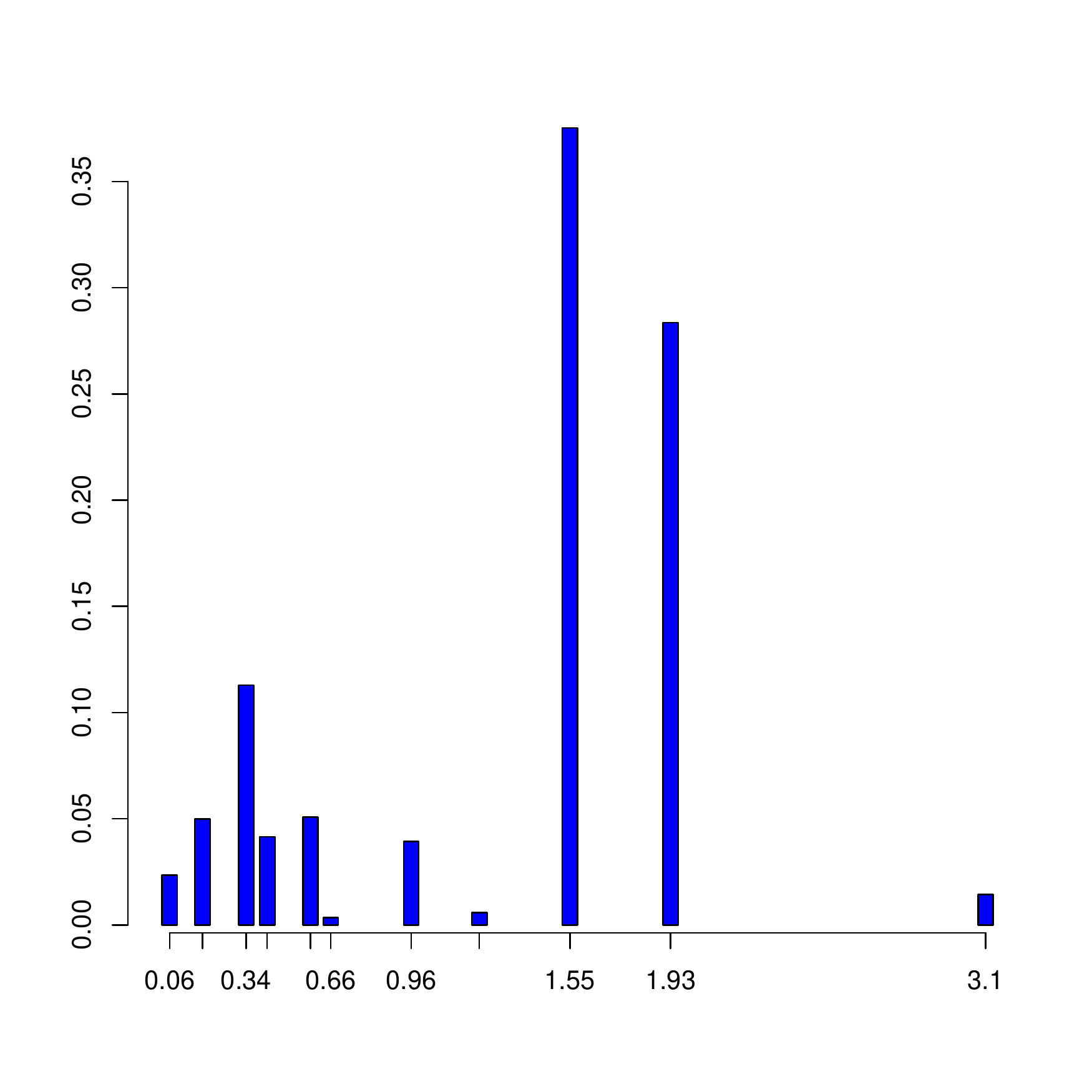}
\includegraphics[width=0.5\textwidth]{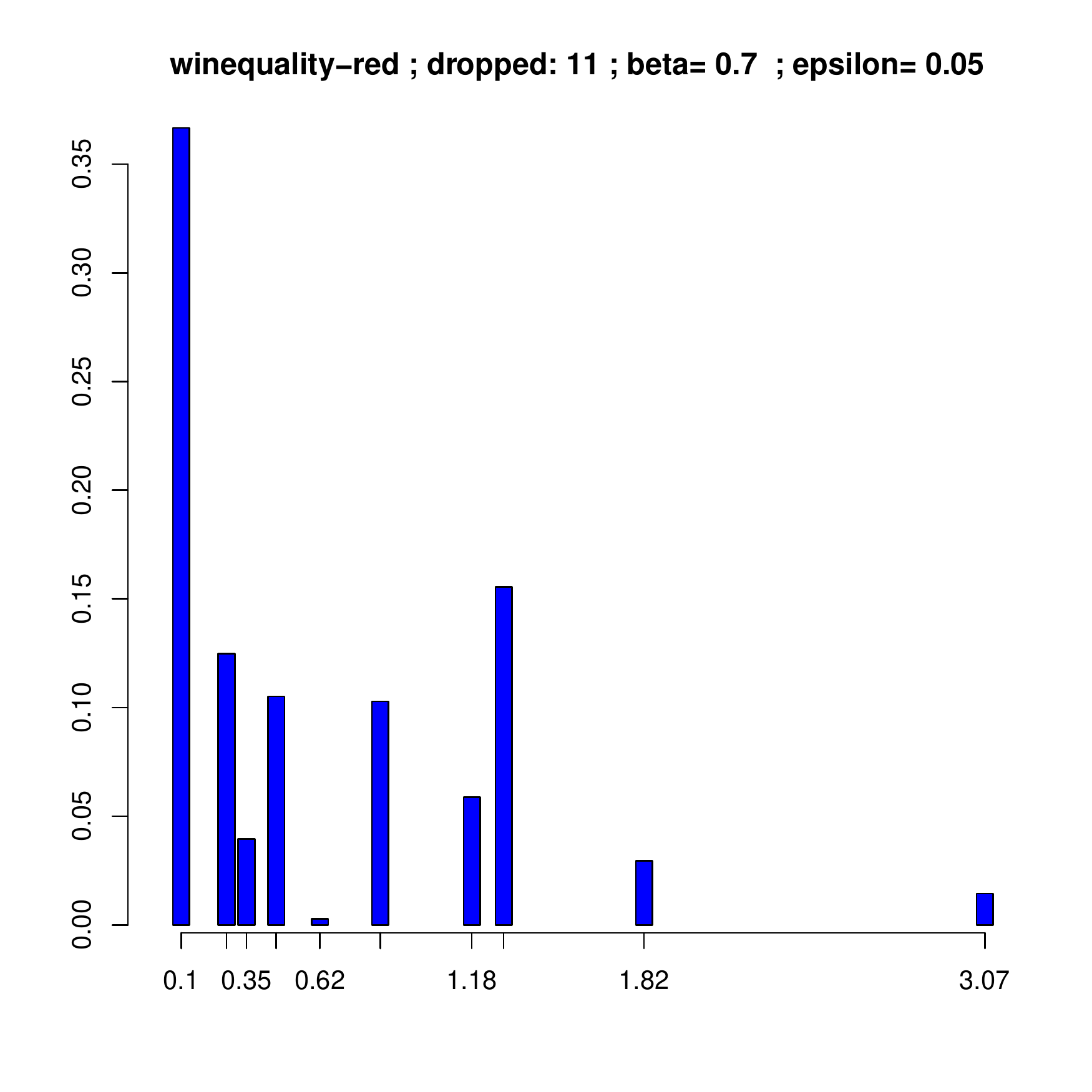}
}
\caption{\label{fig:spectralmeasurewine} 
Spectral measure $\mu_{\sx,\hat{\ba}}$ for the taste of wine for the case
where all ingredients are included (left) versus the case where
$X_{11}$ (alcohol) has been dropped (right). In the latter case, the weights decrease for the larger eigenvalues. } 
\end{figure}
The estimated confounding strength reads $\hat{\beta}=0$.

Since the above experiments suggest that the set of ingrediences influence the target variable taste essentially in an unconfounded way, we now explore 
what happens when we exclude one of the variables
$\bX:=(X_1,\dots,X_{11})$. Since this variable $X_j$
will typically be correlated with the remaining 
ones and since it, at the same time, influences $Y$,
this will typically confound the causal relation
between $\bX\setminus X_j$ and $Y$.
For each $j\in \{1,\dots,11\}$ we have 
therefore estimated the structural confounding strength
and obtained the following results:
for all $j<11$ the algorithm estimated the confounding
strength $0.0$ (the lowest possible value), while it estimated $0.55$ for $j=11$.
Since $X_{11}$ has the strongest influence on $Y$, this result is remarkable because it is plausible that dropping it 
corresponds to strong confounding.

Figure~\ref{fig:spectralmeasurewine}, left visualizes the weights
of the spectral measure for the case where all $11$ variables are included
and compares it to the one obtained when alcohol is dropped (right).
In the latter case, one clearly sees that the weights decrease towards large
eigenvalues, which indicates confounding.



\subsection{Chigaco crime data}

This dataset \cite{chicagocrime} reports\footnote{This site provides applications using data that has been modified for use from its original source, www.cityofchicago.org, the official website of the City of Chicago.  The City of Chicago makes no claims as to the content, accuracy, timeliness, or completeness of any of the data provided at this site.  The data provided at this site is subject to change at any time.  It is understood that the data provided at this site is being used at one’s own risk} the number of crimes
for each of $77$ community areas in Chicago, USA,
and  some potential factors influencing the crime rate
\cite{chicagocrime}.
Here, $Y$ denotes the assaults (homicides)
and $X$ consists of the following $6$ features:
$X_1$: below poverty level, $X_2$: crowded housing,
$X_3$:  dependency, $X_4$: no highschool diploma,
$X_5$: per capita income, $X_6$: unemployment.
After normalization we obtain the following
estimated vector of structure coefficients:
\[
\hat{\ba}=(3.3,3.5,2.8,-7.7,-2.6,9.7)\,.
\] 
It seems reasonable that the unemployment rate 
has the strongest influence. It is, however, surprising
that `no highschool diploma' should have  a negative
influence on the number of crimes. This is probably due to a confounding effect.
The estimated confounding strength reads $\hat{\beta}=0.07$.

\subsection{Compressive strength and ingredients of concrete}

This experiment considers the data set  `concrete and compressive strength' \cite{concrete} in
the machine learning repository.

$Y$ is the compressive strength in megapascals
and $X_1$ to $X_7$ are the following components, measured in $kg/m^3$
$X_1$: cement, $X_2$: blast furnace, $X_3$: fly ash,  $X_4$: water, 
$X_5$: superplasticizer, $X_6$: coarse aggregate, 
$X_7$: fine aggregate. $X_8$ is the age in days.
After normalization, the estimated vector of structure coefficients reads
\[
\hat{\ba}=(12.5,9.0,5.6,-3.2,1.7,1.4,1.6,7.2)
\]
The amount of
superplasticizer seems to have the strongest influence, followed by cement. 
The estimated confounding strength reads
$\hat{\beta}=0.83$, but it is hard to speculate about 
possible confounders here.

\section{Discussion}

We have described a method that estimates the strength of 
a potential one-dimensional common cause $Z$ that confounds the
causal relation between a $d$-dimensional cause  $\bX$ (with `large' $d$) and a one-dimensional effect $Y$. The presence of
$Z$ can, to some extent, be detected from the joint statistics of $\bX$ and $Y$ when the vector $\ba$ of regression coefficients
(after regressing $Y$ on $\bX$) is decomposed into the eigenvectors of the covariance matrix $\sx$  of  $\bX$. 
This is because generically, without confounding, the weights of
this decomposition are roughly uniformly spread over the
principal values of $\bX$, while
the presence of 
$Z$ will typically modify the weights in a way that is characteristic for the corresponding confounding scenario.

The method is based on the assumption that the vector
$\ba$ has, in a certain sense, `generic orientation' with respect to $\sx$. The justification of our method
relies on a highly idealized model where the vectors of model parameters are randomly generated from a rotation invariant 
prior. This yields to several concentration of measure phenomena for high dimensions $d$ which the method employs. 
There is some hope that empirical data show similar concentration of measure phenomena although our model assumptions are probably significantly violated. 

Given the difficulty of the enterprise of inferring causal relations from observational data, one should not expect that any method is able to detect the presence of confounders with certainty. Following this modest attitude, the results can be considered encouraging; after all the joint distribution  of $\bX$ and $Y$ seems to contain some hints on  whether or not their causal relation
is confounded. 

Although the theoretical justification of the method (using asymptotic for dimension to infinity) suggests that the methods should only be applied to large dimension, it should be emphasized 
that we have so far computed the regression without regularization which quickly requires prohibitively high sample sizes. Future work may apply the method following regularized regression but then one has to make sure that the regularizer does not spoil the method by violating our symmetry assumptions.

\vspace{0.2cm}
\noindent
{\bf Acknowledgements:} We would like to thank 
Uli Wannek for helping us with the implementation of
the video experiment. Many thanks also to
Roland Speicher and his group in Saarbr\"ucken for 
helpful discussions about free probability theory
and to Steffen Lauritzen for pointing out that 
real data sometimes show the MTP2 property, which may be an issue here.


\section{Appendix}

\subsection{Relation to free independence}

Free probability theory defines a notion of independence that is asymptotically
satisfied for independently chosen high-dimensional random matrices.
To sketch the idea, we start with a model of generating independent
random matrices considered in \cite{Speicher97}:

Let $(F_d)_{d\in \N}$ and $(G'_d)_{d\in \N}$ be sequences of $d\times d$ matrices
whose tracial spectral measures converge weakly. Let $(U_d)_{d\in \N}$
be random orthogonal matrices drawn from $O(d)$ according to the Haar measure.
We start with the simplest case of the independence conditions:
$F_d$ and $G_d:= U_d G'_d U_d^T$ satisfy asymptotically the Trace Condition, i.e.,
\[
|\tau (F_d G_d)-\tau (F_d) \tau (G_d)|  \to 0
\]
in probability, where $\tau$ denotes again the renormalized trace.  It is then convenient to introduce
limit objects $F,G$ 
 in a $C^*$-algebra \cite{Mu90} and a functional $\phi$, expressing the limit of renormalized traces,
for which the Trace Condition then holds exactly:
\begin{equation}\label{eq:asymptrace}
\phi(FG)=\phi(F)\phi(G)\,.
\end{equation}
However, (\ref{eq:asymptrace}) is only
the simplest one of an infinity of independence statements. 
First, one obtains statements on higher moments
 like 
\[
\phi(F^k G^k)=\phi(F^k)\phi(G^l)\,,
\]
which is analog to $\Exp[X^k Y^l]=\Exp[X^k] \Exp[Y^l]$ for independent random variables $X$ and $Y$.
But the model above also yields independence statements like
$\phi(FGFG) =0$ whenever $\phi(F)=\phi(G)=0$,
which have no counterpart with classical random variables.
$F$ and $G$ are also considered  `non-abelian
random variables' and free independence as a stronger 
version of usual statistical independence which can only hold because the variables do not commute. 

The following difficulty arises when we try to apply the above ideas
to our generating model in Subsection~\ref{subsec:justi}:
Our sequence $(\sx^d)_{d\in \N}$ may take the role of
$(F_d)_{d\in \N}$. To draw $\ba_d$ uniformly from the unit sphere, we may define
an arbitrary sequence $(\ba'_d)$ of unit vectors and set $\ba_d:=U_d \ba_d'$
with $U_d$ being a random rotation as above.  
Then one could naively argue that
$\ba'_d (\ba')^T_d$ takes the role of $G'_d$ and one also expects
\[
\tau ((\sx^d)^k \ba_d \ba_d^T) \approx \tau((\sx^d)^k) \tau(\ba_d \ba_d^T)\,,
\]
which is equivalent to 
\[
\langle \ba_d, (\sx^d)^k \ba_d\rangle \approx \tau((\sx^d)^k) \|\ba_d\|^2\,.
\]
Hence,
\[
\int s^k d\mu_{\sx^d,\ba_d}(s) \approx \int s^k s\mu_{\sx^d,\tau} (s)\,,
\]
for all $k\in \N$ for large $d$. Thus, all moments of $\mu_{\sx,\ba}$ coincide
almost with $\mu_{\sx,\tau}\|\ba\|^2$. 
This argument, however, blurs the fact that $\tau(\ba_d \ba_d)$ converges to zero
while ${\rm tr}(\ba_d \ba_d^T)$ is constant, i.e., we need to consider
the asymptotic of $d \cdot \tau(F_d G_d)$ instead of $\tau(F_d G_d)$, which is not covered by free probability theory to the best of our knowledge.
Theorem~\ref{thm:justi} is thus close to the above statements
although we do not see any straightforward way to derive it from existing work.

\subsection{Correlative versus structural strength of confounding}

To show that the relation between the correlative and structural confounding strength $\gamma$ and $\beta$  is quite sophisticated, we mention the following result:
\begin{Lemma}[correlative vs.
structural strength]\label{lem:corrvsstr}
Let $m_j$ denote the $j$th moment of the tracial spectral measure of $\sx$, i.e., 
\[
m_j:=\tau(\sx^j)  \quad \forall j\in \Z\,.
\]
If Postulate~\ref{post:genorient} holds, $\beta$ and $\gamma$ are related via
the following non-linear functions:
\begin{equation}\label{eq:gammatobeta}
\beta \approx \frac{\gamma  m_{-2} \|\Sigma_{\bX Y}\|^2}{\|\hat{\ba}\|^2 (1+ \gamma m_{-1}  \|\Sigma_{\bX Y}\|^2)^2}\,, 
\end{equation}
and
\begin{equation}\label{eq:betatogamma}
\gamma \approx 
\frac{\left(\sqrt{m_{-2}} + \sqrt{m_{-2}-4 m_{-1} \beta \|\hat{\ba}\|^2 }\right)^2}{4 m^2_{-1} \beta \|\hat{\ba}\|^2 \|\Sigma_{\bX Y}\|^2}\,.
\end{equation}
The $\approx$ signs get a precise meaning by the following statement: left hand sides of (\ref{eq:gammatobeta}) and
(\ref{eq:betatogamma}) converge to the right hand sides
for every sequence of models for which the left hand sides of (\ref{eq:ageneric}), (\ref{eq:bgeneric}), and (\ref{eq:genericvecvec}) converge weakly to the right hand sides. 
\end{Lemma}
{\bf Proof:}
We write all proofs with $\approx$-sign and keep in mind that it means convergence for all sequences of models for which the left hand sides of (\ref{eq:ageneric}), (\ref{eq:bgeneric}), and (\ref{eq:genericvecvec}) converge weakly to the right hand side.
Due to the Sherman-Morrison formula \cite{Sherman49} we have
\[
\sx^{-1} =(\se +\bb \bb^T)^{-1} = \se^{-1}-\frac{\se^{-1} \bb \bb^T \se^{-1}}{1+\langle \bb, \se^{-1} \bb \rangle }\,.
\]
We thus find
\begin{eqnarray*}
\sx^{-1} \bb &=& \se^{-1} \bb \left( 1-  \frac{\langle \bb, \se^{-1} \bb \rangle  }{1+ \langle \bb, \se^{-1} \bb \rangle   } \right)
=
\se^{-1} \bb  \frac{1}{1+\langle \bb, \se^{-1} \bb \rangle} 
\approx   \se^{-1} \bb  \frac{1}{1+ m_{-1} \|\bb\|^2 } \,.
\end{eqnarray*}
Using $\langle \bb,\se^{-2} \bb\rangle \approx
\tau (\se^{-2}) \|\bb\|^2 \approx \tau(\sx^{-2}) \|\bb\|^2 =m_{-2} \|\bb\|^2$ we thus get
\begin{eqnarray}\label{eq:relsxbb}
\|\sx^{-1} \bb\|^2 &\approx &  \frac{m_{-2} \|\bb\|^2 }{(1+m_{-1} \|\bb\|^2)^2} \,.
\end{eqnarray}

Due to $\|\bb\|^2 =\gamma \|\Sigma_{\bX Y}\|^2$ we obtain
\[
\|\sx^{-1} \bb\|^2  \approx \frac{m_{-2} \gamma \|\Sigma_{\bX Y}\|^2}{(1+m_{-1} \gamma \|\Sigma_{\bX Y}\|^2)^2}\,, 
\]
which proves the first part of the statement.

Setting $\theta:=\|\sx^{-1} \bb\|$ and $\alpha:=\|\bb\|$ in
(\ref{eq:relsxbb})  yields
$
\theta \approx \sqrt{m_{-2}} \alpha /(1+m_{-1} \alpha^2)\,.
$
Hence,
$
\theta m_{-1} \alpha^2 -\sqrt{m_{-2}} \alpha +\theta \approx 0\,.
$
This quadratic equation can be solved for $\alpha$ 
whenever 
$
m_{-2} \geq 4 \theta^2 m_{-1} \,,
$
where it yields the unique solution 
\begin{equation}\label{eq:alphasolution}
\alpha \approx\frac{\sqrt{m_{-2}} + \sqrt{m_{-2}-4\theta^2  m_{-1} }}{2 \theta m_{-1}}\,,
\end{equation}
because we need to reject the negative solution.
On the other hand, 
\begin{equation}\label{eq:etatobeta}
\theta=\|\sx^{-1} \bb\| \approx \sqrt{\beta} \|\hat{\ba}\|\,,
\end{equation}
due to (\ref{eq:defstructuralstrength}) and (\ref{eq:betaApp}).  
Inserting (\ref{eq:etatobeta}) into (\ref{eq:alphasolution}) yields
\begin{equation}\label{eq:bnormfrombetaPrime}
\|\bb\| \approx \frac{\sqrt{m_{-2}} + \sqrt{m_{-2}-4 m_{-1} \beta \|\hat{\ba}\|^2 }}{2 m_{-1} \sqrt{\beta} \|\hat{\ba}\|}\,.
\end{equation}
Recalling the definition of $\gamma$ in (\ref{eq:defcorrelativestrength}) and the approximation (\ref{eq:strengthorth}) we obtain
\[
\gamma \approx \frac{\|\bb\|^2}{\|\Sigma_{\bX Y}\|^2} \approx
\frac{\left(\sqrt{m_{-2}} + \sqrt{m_{-2}-4 m_{-1} \beta \|\hat{\ba}\|^2 }\right)^2}{4 m^2_{-1} \beta \|\hat{\ba}\|^2 \|\Sigma_{\bX Y}\|^2}\,.
\]
$\Box$

\end{document}